\documentclass[letterpaper]{article}
\usepackage[preprint]{aaai2027}  % DO NOT CHANGE THIS
\usepackage[hyphens]{url}  % DO NOT CHANGE THIS
\usepackage{graphicx} % DO NOT CHANGE THIS
\usepackage{natbib}  % DO NOT CHANGE THIS AND DO NOT ADD ANY OPTIONS TO IT
\usepackage{caption} % DO NOT CHANGE THIS AND DO NOT ADD ANY OPTIONS TO IT
\usepackage{algorithm}
\usepackage{algorithmic}

\usepackage{newfloat}

\usepackage{listings}
\DeclareCaptionStyle{ruled}{labelfont=normalfont,labelsep=colon,strut=off} % DO NOT CHANGE THIS
\floatstyle{ruled}
\newfloat{listing}{tb}{lst}{}
\floatname{listing}{Listing}

\usepackage{booktabs}
\usepackage{multirow}%%%%
\usepackage{amsmath}%%%%
\usepackage{amssymb}
\usepackage{tabularx}
\usepackage{array}
\title{SW-ProxyCE: Zero-Query Adversarial Transfer from Public EEG Encoders to Private Downstream Models}
\author{
  {Linhua Cong\textsuperscript{1}}\thanks{The first two authors contribute equally.} \quad % \footnotemark[1] 表示与前一个 \thanks共享同一个脚注标记
  {Dingkun Liu\textsuperscript{1}}\footnotemark[1] \quad
  {Dongrui Wu\textsuperscript{1}}\thanks{Corresponding author.} \\ % 第二个 \thanks 命令，用于通讯作者
}
\affiliations{
    \textsuperscript{\rm 1}Huazhong University of Science and Technology\\
    {\{m202476968, liudingkun\}@hust.edu.cn} \\
    {drwu09@gmail.com}
}

\begin{document}

\maketitle

\begin{abstract}

Electroencephalography (EEG) foundation models have recently emerged as a promising paradigm for EEG decoding by learning reusable representations from large-scale heterogeneous neural recordings. However, the open release of EEG foundation encoders, while facilitating downstream developments, also introduces a previously unexplored security risk: publicly available representations may make private downstream models vulnerable. This paper investigates adversarial transfer attacks in EEG foundation model deployment in a public-encoder and private-downstream setting, where attackers have white-box access to a released encoder and a small task-matched labeled reference set, but no access or query to victim parameters, outputs, or gradients. We propose Shrinkage-Whitened Proxy Cross-Entropy (SW-ProxyCE), a query-free task-aware attack framework that recovers task-level decision geometry from a small labeled reference set through shrinkage-whitened class prototypes, enabling transferable adversarial generation without training an additional surrogate classifier. We evaluated SW-ProxyCE across three EEG tasks using three general-purpose foundation encoders and a paradigm-specific pre-trained encoder, covering both linear-probing and full-fine-tuning downstream models in cross-subject and within-subject scenarios. Results demonstrated that adversarial examples generated from the public encoder and limited labeled references can effectively transfer to inaccessible downstream models. SW-ProxyCE consistently outperformed task-agnostic representation-shift attacks, revealing that the strong transferability of EEG foundation models does not necessarily lead to adversarial robustness. Our code will be available on GitHub.\footnote{\url{https://github.com/ccclh/SW-ProxyCE}}

\end{abstract}

\begin{figure}[t]
\centering
\includegraphics[width=\columnwidth]{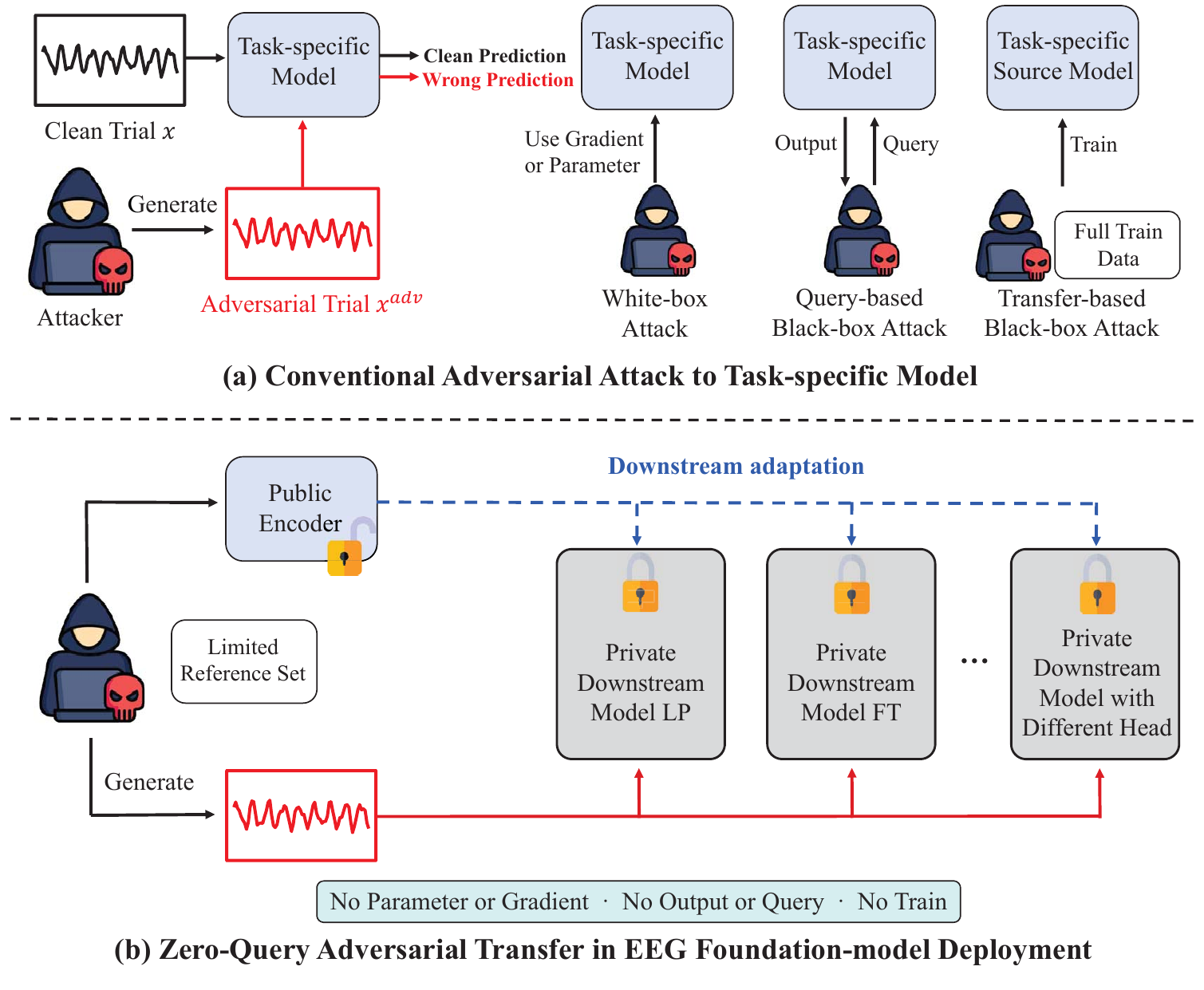}
\caption{Comparison between conventional adversarial attacks on task-specific models and the zero-query adversarial transfer setting from a public EEG encoder to private downstream models.}
\label{fig:Intro}
\end{figure}

\section{Introduction}

Brain--computer interfaces (BCIs) establish a direct communication pathway between neural activity and external devices, enabling users to convey intentions and interact with assistive or intelligent systems without relying on conventional neuromuscular pathways~\cite{cincotti2008non}. Electroencephalography (EEG) is one of the most widely used noninvasive modalities for BCIs, and its practical utility critically depends on accurate decoding of neural dynamics into task-relevant states, intentions, or control commands~\cite{tariq2018eeg}. Conventional EEG decoders, however, are typically developed for specific subjects, datasets, or paradigms and often generalize poorly across subjects and recording configurations~\cite{liu2025sdda}, as well as downstream tasks~\cite{liu2026eeg}. EEG foundation models have recently emerged as a promising approach to improving such generalization. By pre-training on large-scale and heterogeneous EEG corpora, these models aim to learn reusable neural representations that can be adapted to diverse downstream tasks through strategies such as linear probing (LP) and full fine-tuning (FT)~\citep{jiang2024labram,cui2024neurogpt,wang2025cbramod,liu2026eeg}. In practice, pre-trained encoder checkpoints are often publicly released and subsequently reused by multiple downstream users to construct independently developed task-specific EEG systems.

Despite this progress, the adversarial security implications of publicly released EEG foundation encoders remain largely unexplored. As summarized in Fig.~\ref{fig:Intro} (a), existing EEG adversarial attacks primarily target complete task-specific models under white-box, query-based black-box, or conventional transfer-based black-box settings~\citep{zhang2019vulnerability,jiang2019active,liu2021universal}. White-box attacks assume access to the parameters and gradients of the victim model, whereas query-based black-box attacks rely on repeated feedback from the victim to estimate an attack direction or train a substitute model. Transfer-based attacks avoid direct victim access but typically generate perturbations through a separately trained task-specific source model that provides an explicit classification objective. These assumptions do not fully characterize EEG foundation-model deployment, which creates a distinct access asymmetry: the upstream pre-trained encoder is publicly available, whereas downstream models adapted from it may remain private and cannot be queried. 

Although these private models may differ in their classification heads, training data, and adaptation strategies, they share a common pre-trained origin and may retain part of the public encoder's representational and gradient structure. Consequently, perturbations optimized through the released encoder may remain effective after downstream adaptation, making the public encoder a potential shared upstream attack surface across otherwise inaccessible EEG systems. This raises a fundamental question: can an attacker exploit a publicly released EEG encoder to generate adversarial examples that transfer to private downstream models derived from it, without accessing or querying those models?

To bridge this gap, we formulate a public-encoder transfer attack setting illustrated in Fig.~\ref{fig:Intro}(b), in which the attacker has white-box access to a released pre-trained EEG encoder and a small task-matched labeled reference set, but cannot access or query the private downstream model. Under this setting, we propose Shrinkage-Whitened Proxy Cross-Entropy (SW-ProxyCE), a victim-query-free and task-aware attack that constructs class prototypes from the limited reference set and derives a proxy classification objective without training an additional surrogate model. SW-ProxyCE further employs shrinkage whitening to reduce the influence of anisotropic within-class variation and stabilize prototype-based comparisons under limited references. Experiments across three EEG tasks, four pre-trained encoders, LP and FT victims, and both LOSO and within-subject protocols demonstrate that SW-ProxyCE transfers effectively to inaccessible downstream models and consistently outperforms task-agnostic representation-shift attacks.

The main contributions of this work are summarized as follows:

\begin{itemize}
    \item We identify and formalize a previously underexplored security risk in EEG foundation-model deployment: a publicly released pre-trained encoder can serve as a shared upstream attack surface for multiple private downstream models adapted from it.

    \item We propose SW-ProxyCE, a victim-query-free and task-aware adversarial transfer framework that constructs a prototype-based cross-entropy objective from a small labeled reference set. Its shrinkage-whitened proxy geometry captures task-level class competition while mitigating anisotropic and poorly estimated within-class variation.

    \item We conduct a systematic evaluation across three EEG tasks, four publicly available pre-trained encoders, LP and FT downstream victims, and both LOSO and within-subject protocols. The results demonstrate that adversarial examples generated solely through a public encoder can transfer effectively to inaccessible downstream models without victim access or queries.
\end{itemize}

\section{Related Work}
\label{sec:related-work}
\subsection{EEG Foundation Models}
\label{sec:eeg_foundation_models}

EEG foundation models (FMs) have recently emerged as a promising paradigm for learning transferable representations from large-scale heterogeneous EEG corpora. Early studies such as BENDR~\citep{kostas2021bendr} and BIOT~\citep{yang2023biot} demonstrated the feasibility of self-supervised EEG representation learning and cross-dataset transfer. Recent models have further explored diverse pre-training objectives and architectures, including autoregressive token modeling~\citep{cui2024neurogpt}, masked spatial-temporal representation learning~\citep{wang2024eegpt}, and compact masked signal reconstruction~\citep{wang2025cbramod}. Beyond general-purpose pre-training, MIRepNet~\citep{liu2026mirepnet} investigated paradigm-specific pre-training for motor imagery decoding.

Despite these advances, existing EEG FM studies mainly focus on improving downstream accuracy and transferability, while their robustness and security under adversarial perturbations remain largely unexplored~\citep{kuruppu2026eeg,liu2026eeg,dai2026eeg}.

\subsection{Adversarial Attacks on EEG Models}
\label{sec:eeg_adversarial_attacks}

Existing studies on adversarial attacks against EEG decoding systems have predominantly considered supervised task-specific models and are commonly categorized into white-box and black-box attacks according to the attacker's access to the victim model~\citep{wu2023physiological,MENGrobustnessbenchmark}. Under white-box access, early studies demonstrated the vulnerability of EEG classifiers to carefully crafted input perturbations~\citep{zhang2019vulnerability} and subsequently extended adversarial manipulation to EEG regression models and BCI spellers~\citep{meng2019whitebox,zhang2021tiny}. Later work investigated more structured and practically motivated perturbations, including sparse attacks over EEG channels, temporal samples, or brain electrical activity mapping (BEAM) representations~\citep{feng2021saga,yu2023perturbing}, physically constrained attacks~\citep{wang2022physically}, attention-guided universal perturbations~\citep{zhong2025attention}, and adversarial filtering-based attacks~\citep{meng2024filtering}.

Black-box EEG attacks generally follow either query-based or transfer-based strategies. Query-based methods exploit feedback from the victim model to train substitute models~\citep{jiang2019active} or iteratively construct low-distortion perturbations using a limited number of hard-label queries~\citep{10974455}. Transfer-based methods instead generate adversarial examples through a task-specific source model and evaluate whether the resulting perturbations remain effective against independently trained target models, typically across different model architectures~\citep{liu2021universal,jung2023generative,yi2024timefrequency}. Although such methods avoid direct access to the target parameters, they generally rely on a complete task-aware source predictor and often assume that the source and target models are trained on the same or closely matched data distributions.

Despite substantial progress, existing EEG adversarial studies largely focus on complete task-specific models and do not capture the public-encoder/private-downstream access asymmetry of EEG foundation-model deployment. To the best of our knowledge, this is the first work to systematically investigate adversarial transfer from publicly released EEG encoders to inaccessible downstream models.

\section{Problem Definition and Security Analysis}
\subsection{Public-Encoder Deployment and Threat Setting}
\label{sec:threat_setting}
In the deployment pattern considered here, a pre-trained EEG encoder is publicly released for downstream reuse, while models adapted from it remain private. Let \(f_{\theta_0}\) denote the public encoder. A private downstream model derived from it is represented as
\begin{equation}
F_{\mathrm{vic}}=\mathcal{A}\left(f_{\theta_0},\mathcal{D}_{\mathrm{task}}\right),
\label{eq:victim_model}
\end{equation}
where \(\mathcal{A}\) represents the downstream adaptation procedure and \(\mathcal{D}_{\mathrm{task}}\) denotes the private task-specific adaptation data. The attacker has white-box access to the public encoder \(f_{\theta_0}\), including its architecture and parameters, and can compute input gradients through this encoder. In contrast, the architecture, parameters, outputs, and gradients of \(F_{\mathrm{vic}}\) are unavailable to the attacker. Moreover, no queries to the victim model are permitted during adversarial example generation.

We assume that the attacker possesses a small task-matched labeled reference set obtained independently of the private downstream model and its training data. Here, task-matched indicates that the reference samples share the same prediction task and label space as the victim model. Such references may be obtained from publicly available datasets with compatible task definitions or collected under a compatible experimental protocol. We denote the reference set as
\begin{equation}
\mathcal{D}_{\mathrm{ref}}=\left\{\left(x_i^{r},y_i^{r}\right)\right\}_{i=1}^{N},\qquad N=KC,
\label{eq:reference_set}
\end{equation}
where \(C\) is the number of task classes and \(K\) is the number of reference samples per class. The reference set is disjoint from both \(\mathcal{D}_{\mathrm{task}}\) and the trials used for attack evaluation.

We consider untargeted perturbations in the digital domain and assume that the attacker can modify an EEG trial before downstream inference. Given a trial \(x\) with label \(y\), the attacker constructs \(x_{\mathrm{adv}}=x+\delta\) subject to \(\|\delta\|_{\infty}\leq\epsilon\). Attack generation uses \(\mathcal{D}_{\mathrm{ref}}\) together with forward passes and input gradients through \(f_{\theta_0}\), without involving \(F_{\mathrm{vic}}\).

\subsection{Downstream Transfer Risk from Public Encoders}
\label{sec:transfer_risk}

Although the downstream model \(F_{\mathrm{vic}}\) remains private, the public encoder \(f_{\theta_0}\) is fully exposed to potential attackers. Since downstream adaptation may preserve part of the representation structure inherited from the pre-trained encoder, adversarial perturbations generated through \(f_{\theta_0}\) may remain effective after adaptation and transfer to private downstream models.

For a trial-label pair \((x,y)\), we characterize the transfer risk induced by the public encoder as
\begin{equation}
\mathcal{R}_{\mathrm{transfer}}(x,y)=\mathcal{E}\left(F_{\mathrm{vic}},\mathcal{G}\left(x,y;f_{\theta_0},\mathcal{D}_{\mathrm{ref}}\right),y\right),
\label{eq:transfer_risk}
\end{equation}
where \(\mathcal{G}(x,y;f_{\theta_0},\mathcal{D}_{\mathrm{ref}})=x_{\mathrm{adv}}\) denotes adversarial-example generation and \(\mathcal{E}\) quantifies its effect on \(F_{\mathrm{vic}}\). At the sample level, \(\mathcal{E}\) is instantiated as \(\mathbb{I}[F_{\mathrm{vic}}(x_{\mathrm{adv}})\neq y]\); over the evaluation set, we assess transfer using balanced accuracy under attack.

Successful transfer demonstrates that keeping downstream parameters private does not necessarily prevent attacks originating from the publicly released encoder. Since a single pre-trained encoder may support multiple independently adapted downstream systems, such vulnerability represents a shared upstream security risk rather than an isolated weakness of an individual model.

\section{Method} \label{sec:method}
\subsection{Overview} \label{sec:method_overview}

Under the public-encoder transfer attack setting, we propose Shrinkage-Whitened Proxy Cross-Entropy (SW-ProxyCE), a victim-query-free adversarial transfer framework that generates perturbations using only the publicly available encoder and a small task-matched labeled reference set. The key challenge is that the public encoder provides discriminative representations but does not expose the task-specific decision function of private downstream models. To recover task-related decision geometry without training an additional surrogate classifier, SW-ProxyCE constructs a prototype-based proxy objective from the labeled reference samples. Furthermore, EEG representations exhibit substantial variability across trials, recording sessions, and subjects, which can manifest as anisotropic within-class dispersion in the encoder space. Under limited references, such variability can distort prototype estimation and compromise the reliability of similarity-based comparison. SW-ProxyCE therefore introduces a shrinkage-whitened proxy geometry to calibrate within-class variation and improve the stability of prototype-based adversarial generation.

\begin{figure*}[t]
\centering
\includegraphics[width=\textwidth]{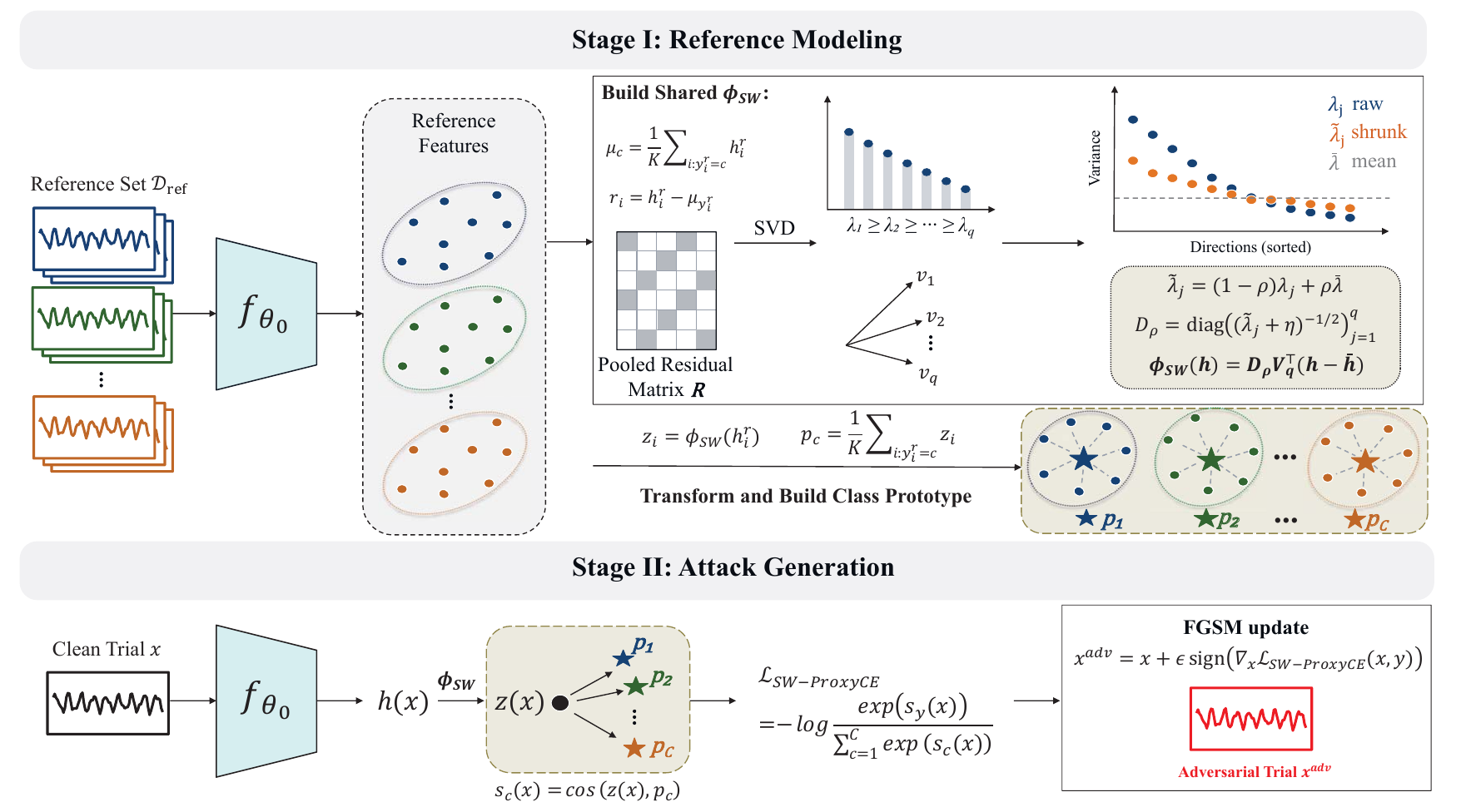}
\caption{Overview of the proposed SW-ProxyCE framework. SW-ProxyCE leverages a public EEG encoder and a small labeled reference set to construct shrinkage-whitened class prototypes and a task-aware ProxyCE objective. The generated adversarial perturbations are optimized through the public encoder and transferred to private downstream models without accessing or querying the victim models.}
\label{fig:method}
\end{figure*}

As illustrated in Fig.~\ref{fig:method}, SW-ProxyCE consists of two stages. In the reference modeling stage, a shared shrinkage-whitened mapping is estimated from class-centered reference representations, and class prototypes are constructed in the resulting calibrated feature space. In the adversarial generation stage, the proposed ProxyCE objective is maximized with respect to the input EEG signal under an \(\ell_\infty\)-bounded perturbation constraint. The private downstream model is not involved in either stage.

\subsection{Task-Aware ProxyCE}
\label{sec:proxyce}
Although the public encoder provides differentiable representations, it does not contain a task-specific decision boundary required for conventional classification-based attacks. We therefore construct a non-parametric task-aware proxy using a small labeled reference set. For each reference trial \(x_i^r\), its encoder representation is obtained as
\begin{equation}
h_i=f_{\theta_0}(x_i^r)\in\mathbb{R}^{d},
\end{equation}
where \(f_{\theta_0}\) denotes the public encoder. A shared feature mapping \(\phi(\cdot)\) is then applied to project encoder representations into a proxy space:
\begin{equation}
z_i=\phi(h_i).
\end{equation}
The same mapping is applied to both reference samples and query samples. For the original ProxyCE formulation, \(\phi\) is set as the identity mapping, i.e., \(\phi_{\mathrm{raw}}(h)=h\), whereas SW-ProxyCE adopts the shrinkage-whitened mapping \(\phi_{\mathrm{SW}}\) introduced in the following subsection.

Given the transformed reference representations, the prototype of class \(c\) is computed as
\begin{equation}
p_c=\frac{1}{K}\sum_{i:y_i^r=c}z_i,\qquad c=1,\ldots,C.
\label{eq:class_prototype}
\end{equation}

For an input trial \(x\), we similarly define its encoder representation \(h(x)=f_{\theta_0}(x)\) and proxy-space representation \(z(x)=\phi(h(x))\). Its proxy logit for class \(c\) is defined by cosine similarity:
\begin{equation}
s_c(x)=\cos\left(z(x),p_c\right).
\label{eq:proxy_logit}
\end{equation}
The resulting ProxyCE loss is defined as
\begin{equation}
\mathcal{L}_{\mathrm{ProxyCE}}(x,y)=-\log\frac{\exp(s_y(x))}{\sum_{c=1}^{C}\exp(s_c(x))}.
\label{eq:proxyce}
\end{equation}
The prototype set together with the query-to-prototype similarities defines the proxy geometry underlying ProxyCE. Maximizing \(\mathcal{L}_{\mathrm{ProxyCE}}\) reduces the relative probability assigned to the ground-truth class while increasing the competition from alternative classes. Since the objective depends only on the public encoder and the reference-derived proxy geometry, its gradient with respect to the input can be computed without accessing or querying the private downstream model. This enables adversarial example generation under the proposed public-encoder transfer setting.

\subsection{Shrinkage-Whitened Proxy Geometry} \label{sec:shrinkage_geometry}

Although the prototype-based ProxyCE objective introduces task-level class information, its reliability depends on the quality of the reference-derived prototype geometry. EEG representations often exhibit substantial intra-class variability across trials, recording sessions, and subjects, even under the same task condition. Such variability is typically anisotropic, where different representation directions exhibit different levels of within-class dispersion (as illustrated in Fig.~\ref{fig:variance_cvr}; see the Appendix for complete results). When only a limited number of reference samples are available, directions dominated by large within-class variation may introduce unstable prototype estimation and disproportionately affect query-to-prototype similarity comparisons. To improve the reliability of the proxy geometry, we construct a shrinkage-whitened projection space that calibrates within-class variation before prototype construction.

We first characterize within-class variability by removing class-specific mean differences from the reference representations. For each class \(c\), the class mean is computed as
\begin{equation}
\mu_c=\frac{1}{K}\sum_{i:y_i^r=c}h_i,\qquad r_i=h_i-\mu_{y_i^r},
\label{eq:class_mean_residual}
\end{equation}
where \(r_i\) represents the deviation of a reference representation from its corresponding class center. These centered residuals remove inter-class differences and isolate variations arising within each class. We stack all residuals into a matrix
\begin{equation}
R=[r_1^\top;\ldots;r_N^\top]\in\mathbb{R}^{N\times d},
\end{equation}
where \(d\) is the encoder feature dimension. Since the available references are limited, residuals from all classes are pooled to estimate a shared within-class variation structure.

Since the residuals within each class sum to zero, the residual matrix satisfies \(\operatorname{rank}(R)\leq N-C\). Let \(q=\operatorname{rank}(R)\). We use the compact SVD to decompose all empirically observed within-class variation:
\begin{equation}
\frac{1}{\sqrt{N-C}}R=U_q\Sigma_qV_q^\top,
\label{eq:within_class_svd}
\end{equation}
where \(\Sigma_q=\operatorname{diag}(\sigma_1,\ldots,\sigma_q)\), and the columns of \(V_q=[v_1,\ldots,v_q]\) span the empirically observed within-class variation subspace. The corresponding directional variances are defined as
\(\lambda_j=\sigma_j^2\).

Direct whitening based on the empirical variances can excessively amplify directions whose variances are inaccurately estimated from limited references. We therefore introduce shrinkage regularization by pulling each directional variance toward the average variance:
\begin{equation}
\widetilde{\lambda}_j=(1-\rho)\lambda_j+\rho\bar{\lambda},
\qquad
\rho\in[0,1],
\label{eq:variance_shrinkage}
\end{equation}
where
\begin{equation}
\bar{\lambda}
=
\frac{\|R\|_F^2}{(N-C)q}.
\end{equation}
The shrinkage coefficient \(\rho\) controls the trade-off between preserving observed directional differences and improving estimation stability.

Let \(\bar{h}=N^{-1}\sum_{i=1}^{N}h_i\) denote the global mean of the reference representations. Centering by \(\bar{h}\) expresses all representations relative to the common reference-set mean before projection and prototype comparison. We further define the shrinkage-whitened projection mapping as
\begin{equation}
\phi_{\mathrm{SW}}(h)
=
D_\rho V_q^\top(h-\bar h),
\label{eq:shrinkage_whitened_mapping}
\end{equation}
where \[
D_\rho=
\operatorname{diag}
((\widetilde{\lambda}_1+\eta)^{-1/2},
\ldots,
(\widetilde{\lambda}_q+\eta)^{-1/2})
\]
and \(\eta>0\) is a numerical stability constant. The projection \(V_q^\top(h-\bar h)\) restricts representations to the empirically supported within-class subspace, while \(D_\rho\) rescales each direction according to its shrinkage-regularized variance. Applying the same mapping to reference and query representations yields a calibrated proxy space where prototype construction and similarity comparison are performed consistently.

Substituting Eq.~\eqref{eq:shrinkage_whitened_mapping} into Eqs.~\eqref{eq:class_prototype}--\eqref{eq:proxyce} yields SW-ProxyCE. The mapping and class prototypes are estimated once from the reference set and remain fixed throughout adversarial example generation.

\subsection{Attack Generation}  \label{sec:attack_generation}
We denote the resulting ProxyCE objective with \(\phi=\phi_{\mathrm{SW}}\) as \(\mathcal{L}_{\mathrm{SW\text{-}ProxyCE}}\). Given a labeled EEG trial \((x,y)\), we generate an adversarial example by maximizing this objective under an \(\ell_\infty\)-bounded perturbation constraint. Specifically, we adopt the Fast Gradient Sign Method (FGSM)~\citep{goodfellow2015explaining} update:
\begin{equation}
x^{\mathrm{adv}}
=
x+
\epsilon\,
\operatorname{sign}
\left(
\nabla_x
\mathcal{L}_{\mathrm{SW\text{-}ProxyCE}}(x,y)
\right),
\label{eq:attack_generation}
\end{equation}
where \(\epsilon\) controls the maximum perturbation magnitude.

During attack generation, the shrinkage-whitened mapping \(\phi_{\mathrm{SW}}\) and the reference-derived class prototypes remain fixed, as they are estimated only once during the reference modeling stage. The gradient is propagated exclusively through the public encoder \(f_{\theta_0}\) and the fixed proxy geometry, without accessing, querying, or relying on any information from the private downstream model \(F_{\mathrm{vic}}\).

\section{Experiments}
\label{sec:experiments}

\subsection{Experimental Setup}
\label{sec:experimental_setup}

\paragraph{Datasets and protocols.}
We evaluated SW-ProxyCE on three EEG decoding tasks: four-class motor imagery on BNCI2014001~\citep{BNCI2014001}, three-class emotion recognition on SEED~\citep{seed}, and binary seizure detection on CHB-MIT~\citep{CHB-MIT}. We considered both leave-one-subject-out (LOSO) and within-subject evaluation. Details of preprocessing and data splits are provided in the Appendix. 

\paragraph{Models and attack settings.}
We evaluated three general-purpose EEG foundation encoders, LaBraM~\citep{jiang2024labram}, Neuro-GPT~\citep{cui2024neurogpt}, and CBraMod~\citep{wang2025cbramod}, together with MIRepNet~\citep{liu2026mirepnet}, a paradigm-specific pre-trained encoder for motor imagery. Private downstream victims were constructed using linear probing (LP) and full fine-tuning (FT). SW-ProxyCE used \(K=20\) reference samples per class, randomly sampled from the development pool before victim training and excluded from the victim-training split, with no overlap with the test set, and generated untargeted adversarial examples using one-step FGSM with \(\epsilon=0.1\) in the z-score-normalized input space. The shrinkage coefficient and stability constant were set to \(\rho=0.4\) and \(\eta=10^{-5}\). Additional analyses of reference sampling, hyperparameter sensitivity, perturbation budgets, and attack optimizers, including MI~\citep{MI} and PGD~\citep{PGD}, are provided in the Appendix.

\paragraph{Baselines and metrics.}
We compared SW-ProxyCE with Task-Agnostic Attack (TAA)~\citep{TAA} and Gaussian noise. We reported Clean BACC, Attacked BACC, and BACC drop, where the latter was defined as \(\mathrm{Clean\ BACC}-\mathrm{Attacked\ BACC}\). Lower Attacked BACC and larger BACC drop indicate stronger attacks. Complementary results using ACC and ASR are reported in the Appendix.

\subsection{Main Results}
\label{sec:main_results}

\begin{table*}[t]
\centering
\begin{tabular*}{\textwidth}{@{}ll@{\extracolsep{\fill}}cccccccc@{}}
\toprule
\multirow{3}{*}{Dataset}
& \multirow{3}{*}{Model}
& \multicolumn{4}{c}{\textbf{LOSO}}
& \multicolumn{4}{c}{\textbf{Within-Subject}}\\
\cmidrule(lr){3-6}\cmidrule(lr){7-10}
& & \multirow{2}{*}{Clean BACC}
& \multicolumn{3}{c}{Attacked BACC \(\downarrow\)}
& \multirow{2}{*}{Clean BACC}
& \multicolumn{3}{c}{Attacked BACC \(\downarrow\)}\\
\cmidrule(lr){4-6}\cmidrule(lr){8-10}
& & & GN & TAA & Ours & & GN & TAA & Ours\\
\midrule
\multicolumn{10}{@{}l@{}}{\textbf{Linear Probing (LP)}}\\
\addlinespace[2pt]
\multirow{4}{*}{BNCI2014001}
& LaBraM & 47.96 & 47.55 & 43.92 & \textbf{23.61} & 54.37 & 54.37 & 46.03 & \textbf{2.18}\\
& Neuro-GPT & 41.22 & 40.91 & 37.23 & \textbf{15.57} & 51.59 & 50.79 & 36.90 & \textbf{11.11}\\
& CBraMod & 45.37 & 45.33 & 41.44 & \textbf{13.45} & 50.20 & 50.00 & 48.41 & \textbf{1.39}\\
& MIRepNet & 49.69 & 49.58 & 30.32 & \textbf{5.63} & 56.94 & 55.01 & 31.60 & \textbf{9.42}\\
\cmidrule{1-10}
\multirow{3}{*}{SEED}
& LaBraM & 53.87 & 46.11 & 35.92 & \textbf{33.37} & 78.25 & 60.96 & 40.48 & \textbf{29.47}\\
& Neuro-GPT & 45.96 & 45.90 & 44.03 & \textbf{23.75} & 50.12 & 50.09 & 46.95 & \textbf{27.41}\\
& CBraMod & 51.71 & 51.66 & 50.45 & \textbf{35.99} & 74.73 & 74.43 & 69.75 & \textbf{24.88}\\
\cmidrule{1-10}
\multirow{3}{*}{CHB-MIT}
& LaBraM & 55.64 & 56.78 & 56.08 & \textbf{31.15} & 70.11 & 62.32 & 59.08 & \textbf{50.45}\\
& Neuro-GPT & 65.36 & 65.49 & 62.27 & \textbf{7.21} & 70.46 & 70.27 & 68.33 & \textbf{39.79}\\
& CBraMod & 54.00 & 53.79 & 51.81 & \textbf{33.78} & 70.52 & 70.37 & 68.62 & \textbf{54.93}\\
\midrule
\multicolumn{10}{@{}l@{}}{\textbf{Full Fine-Tuning (FT)}}\\
\addlinespace[2pt]
\multirow{4}{*}{BNCI2014001}
& LaBraM & 52.28 & 51.99 & 51.22 & \textbf{36.96} & 49.80 & 49.01 & 42.46 & \textbf{4.56}\\
& Neuro-GPT & 54.88 & 54.46 & 50.15 & \textbf{38.58} & 50.20 & 50.00 & 41.27 & \textbf{25.99}\\
& CBraMod & 53.65 & 53.74 & 53.74 & \textbf{46.82} & 56.35 & 55.95 & 51.98 & \textbf{49.01}\\
& MIRepNet & 58.33 & 58.06 & 33.06 & \textbf{9.72} & 70.49 & 70.44 & 31.45 & \textbf{10.57}\\
\cmidrule{1-10}
\multirow{3}{*}{SEED}
& LaBraM & 55.37 & 51.07 & 49.67 & \textbf{47.67} & 86.04 & 71.23 & 46.49 & \textbf{44.93}\\
& Neuro-GPT & 53.40 & 53.33 & 53.17 & \textbf{48.31} & 73.65 & 73.54 & 71.04 & \textbf{64.10}\\
& CBraMod & 52.71 & 52.31 & 48.85 & \textbf{44.98} & 87.00 & 85.59 & 81.63 & \textbf{72.46}\\
\cmidrule{1-10}
\multirow{3}{*}{CHB-MIT}
& LaBraM & 58.95 & 59.16 & 59.02 & \textbf{48.69} & 64.02 & 63.49 & 62.57 & \textbf{60.62}\\
& Neuro-GPT & 68.46 & 68.38 & 67.38 & \textbf{62.86} & 85.51 & 85.61 & 82.90 & \textbf{63.40}\\
& CBraMod & 58.80 & 58.64 & 58.27 & \textbf{53.16} & 81.80 & 81.73 & 81.48 & \textbf{77.37}\\
\bottomrule
\end{tabular*}
\caption{Transfer attack performance against private downstream models under LOSO and within-subject evaluation. Results are reported as BACC (\%). GN denotes Gaussian noise, TAA denotes Task-Agnostic Attack, and Ours denotes SW-ProxyCE. Lower Attacked BACC indicates stronger attacks.}
\label{tab:main_results}
\end{table*}

Table~\ref{tab:main_results} summarizes the transfer attack performance across 40 configurations. SW-ProxyCE consistently achieves the lowest Attacked BACC among all compared methods. Across the 36 configurations involving general-purpose encoders, SW-ProxyCE produces an average BACC drop of 22.90 percentage points, substantially exceeding TAA (5.93 points). These results demonstrate the effectiveness of the proposed task-aware proxy objective for generating transferable adversarial examples from public EEG encoders.

SW-ProxyCE remains effective against both LP and FT downstream victims, achieving average BACC drops of 31.78 and 14.02 percentage points, respectively. The stronger degradation observed on LP models is expected because they directly preserve the public encoder representation, whereas FT introduces representation changes that partially reduce transferability. Notably, MIRepNet exhibits larger performance degradation (44.06--59.92 percentage points), suggesting that paradigm-specific pre-training may preserve more task-relevant structures and provide a more informative proxy geometry for adversarial transfer. 

\subsection{Analysis of SW-ProxyCE} \label{sec:method_analysis}

\paragraph{Effectiveness of the task-aware proxy objective.}
To evaluate the contribution of the proposed task-aware objective, we compared ProxyCE with two alternative attack objectives using the same reference information: Nearest-Competitor Shift (NC-Shift), which optimizes the representation toward the closest competing class, and Reference-Surrogate FGSM (Ref-Surrogate), which trains a surrogate classifier on the reference set. Unlike these approaches, ProxyCE directly models the competition among all classes through prototype-based scores without introducing an additional classifier.

As shown in Table~\ref{tab:proxy_objective}, ProxyCE achieves the largest BACC drop across all three encoders for both LP and FT victims. Its improvement over NC-Shift indicates that considering the global competition among all classes is more effective than optimizing toward a single competing class. Compared with Ref-Surrogate, ProxyCE achieves stronger transferability under limited reference data, demonstrating the advantage of non-parametric prototype-based task modeling.

\begin{table}[t]
\centering
\begin{tabular*}{\columnwidth}{@{\extracolsep{\fill}}lccc@{}}
\toprule
& \multicolumn{3}{c}{BACC Drop \(\uparrow\) (LP / FT)}\\
\cmidrule(lr){2-4}
Attack Proxy & LaBraM & Neuro-GPT & CBraMod\\
\midrule
NC-Shift
& 5.13 / 1.76
& 8.58 / 5.40
& 11.36 / 0.93\\
Ref-Surrogate
& 15.61 / 10.11
& 12.38 / 7.18
& 4.26 / 2.89\\
ProxyCE
& \textbf{18.58 / 12.31}
& \textbf{15.63 / 8.10}
& \textbf{23.73 / 3.68}\\
\bottomrule
\end{tabular*}
\caption{Comparison of task-aware proxy objectives on BNCI2014001 under LOSO evaluation. Values denote BACC drop (\%) for LP and FT victims. 
}
\label{tab:proxy_objective}
\end{table}

\paragraph{Effectiveness of shrinkage-whitened geometry.}

\begin{figure}[t]
\centering
\includegraphics[width=\columnwidth]{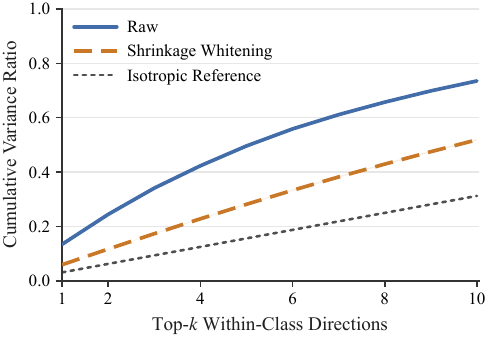}
\caption{Cumulative within-class variance ratio for Neuro-GPT on BNCI2014001 with \(K=20\) references per class. 
}
\label{fig:variance_cvr}
\end{figure}

We further investigated whether calibrating the proxy geometry improves attack transferability. Figure~\ref{fig:variance_cvr} visualizes the cumulative within-class variance ratio (CVR) of Neuro-GPT representations. The raw representation space exhibits highly concentrated within-class variation, whereas shrinkage whitening produces a more balanced variance distribution by reducing the dominance of leading directions.

To quantify the impact of this geometric calibration, we compared three variants: ProxyCE in the raw encoder space, W-ProxyCE with ordinary whitening, and SW-ProxyCE with shrinkage whitening. All variants used the same reference set and attack objective. As reported in Table~\ref{tab:geometry_ablation}, SW-ProxyCE consistently achieves the strongest attack performance across all evaluated encoders, validating the effectiveness of shrinkage regularization under limited reference samples.

\begin{table}[t]
\centering
\begin{tabular*}{\columnwidth}{@{\extracolsep{\fill}}lccc@{}}
\toprule
& \multicolumn{3}{c}{BACC Drop \(\uparrow\) (LP/FT)}\\
\cmidrule(lr){2-4}
Proxy Geometry & LaBraM & Neuro-GPT & CBraMod\\
\midrule
ProxyCE
& 18.58/12.31
& 15.63/8.10
& 23.73/3.68\\
W-ProxyCE
& 9.80/6.27
& 18.60/11.07
& 31.38/4.61\\
SW-ProxyCE
& \textbf{24.35/15.32}
& \textbf{25.65/16.30}
& \textbf{31.92/6.83}\\
\bottomrule
\end{tabular*}
\caption{Ablation of proxy geometry on BNCI2014001 under LOSO evaluation. Values denote BACC drop (\%) for LP and FT victims.}
\label{tab:geometry_ablation}
\end{table}

\paragraph{Effectiveness of reference-set size.}

\begin{figure}[t]
    \centering
    \includegraphics[width=\columnwidth]{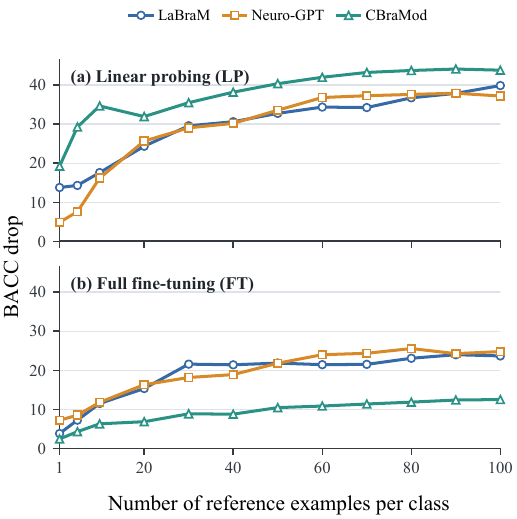}
    \caption{Effect of the per-class reference-set size \(K\) on BNCI2014001 under LOSO evaluation. Panels (a) and (b) show LP and FT victims.}
    \label{fig:reference_size}
\end{figure}

We examined the influence of reference-set size by varying the number of labeled samples per class \(K\) from 1 to 100 on BNCI2014001 under LOSO evaluation. As shown in Figure~\ref{fig:reference_size}, attack effectiveness generally increases with additional reference samples. When \(K=1\), shrinkage whitening is unavailable due to insufficient samples for estimating within-class variation, and Raw ProxyCE is used instead. Even in this extreme low-reference setting, the attack remains effective, reducing BACC by 2.41--19.25 percentage points across the evaluated victims. With increasing \(K\), the average BACC drop improves from 20.06 percentage points at \(K=20\) to 30.31 percentage points at \(K=100\). These results demonstrate that public EEG encoders can expose downstream models even with limited task-matched references, while richer reference information further strengthens the transfer risk.

\section{Conclusion}
\label{sec:conclusion}

This paper investigated the adversarial security risk introduced by publicly released EEG foundation encoders under a public-encoder and private-downstream deployment setting. We showed that a publicly available encoder can serve as an effective source for transferable adversarial attacks against downstream models adapted from it, even when the victim models remain inaccessible and unqueryable. To address this problem, we proposed SW-ProxyCE, a victim-query-free adversarial transfer framework that constructs a task-aware proxy objective from reference-derived class prototypes and shrinkage-whitened representation geometry, without requiring an additional surrogate classifier. Extensive experiments across diverse EEG tasks, foundation encoders, and downstream adaptation strategies demonstrated that adversarial examples generated through public encoders can effectively transfer to private downstream models and consistently outperform task-agnostic attack strategies. These findings reveal that the representation transferability enabled by EEG foundation models does not inherently guarantee adversarial robustness, highlighting the importance of considering encoder-level security risks when developing and deploying publicly accessible EEG foundation models.

\bibliography{aaai2027}


% % \documentclass[letterpaper]{article}
% % \usepackage[submission]{aaai2027}
% % \usepackage[hyphens]{url}
% % \usepackage{graphicx}
% % \urlstyle{rm}
% % \def\UrlFont{\rm}
% % \usepackage{natbib}
% % \usepackage{caption}
% % \frenchspacing

% % \usepackage{booktabs}
% % \usepackage{multirow}
% % \usepackage{amsmath}
% % \usepackage{amssymb}
% % \usepackage{tabularx}
% % \usepackage{array}

% \pdfinfo{
% /TemplateVersion (2027.1)
% }

% \setcounter{secnumdepth}{0}

% \title{Supplementary Material for\\
% SW-ProxyCE: Zero-Query Adversarial Transfer from Public EEG Encoders to Private Downstream Models}

% \author{Anonymous Authors}
% \affiliations{}

% \begin{document}

% \maketitle

% \bibliography{aaai2027}

% \end{document}

\clearpage

\section{A. Dataset Preprocessing and Data Split Details}
\label{app:dataset_preprocessing}
This section provides detailed information on dataset construction, preprocessing, and evaluation splits for all downstream tasks.

\paragraph{BNCI2014001.}
BNCI2014001, also known as BCI Competition IV Dataset 2a~\citep{BNCI2014001}, contains four-class motor imagery EEG recordings from nine subjects. Each subject completed one training session and one evaluation session, each containing 288 trials with 72 trials per class. Thus, each subject contributed 576 trials, including 144 trials for each of the left-hand, right-hand, both-feet, and tongue classes. The 22 EEG channels were sampled at 250 Hz, epoched from 0 to 4 s after each cue, band-pass filtered to 0-38 Hz, scaled to microvolts, and processed using exponential moving standardization. Before model input, each trial was common-average centered, temporally interpolated to 800 samples for LaBraM and CBraMod or 1000 samples for NeuroGPT, and standardized by temporal z-score independently for each channel. For MIRepNet, an MI-specific foundation model, we used a separate preprocessing pipeline: the raw signals were filtered to 8--30 Hz, scaled to microvolts, truncated to 1000 time samples, processed using Euclidean Alignment, mapped from 22 to 45 channels by inverse-distance interpolation, and standardized by temporal z-score independently for each channel.

\paragraph{SEED.}
SEED~\citep{seed} contains three-class emotion EEG recordings from 15 subjects, each recorded over three sessions. Each session contains 15 film trials, with five trials for each of the negative, neutral, and positive classes. We utilized the raw CNT recordings and selected the canonical 62 EEG channels. The signals were low-pass filtered at 75 Hz, resampled to 200 Hz, and segmented into non-overlapping 1-s windows, producing 152,730 windows in total. Each subject contributed 10,182 windows, including 3,360 negative, 3,312 neutral, and 3,510 positive windows. Each channel was independently standardized by temporal z-score before model input; NeuroGPT additionally interpolated each window to 400 samples and divided it into two 200-sample chunks.

\paragraph{CHB-MIT.}
CHB-MIT~\citep{CHB-MIT} was used for binary seizure detection. The processed dataset contains 23 subjects and 29,840 windows, including 27,150 non-seizure and 2,690 seizure windows. Ictal and adjacent non-ictal signals from 18 bipolar channels were segmented into non-overlapping 4-s windows. Each window was standardized by temporal z-score independently for each channel. LaBraM used 200 Hz signals filtered at 0.1--75 Hz with a 50 Hz notch; NeuroGPT used 250 Hz signals filtered at 0.5--100 Hz with a 60 Hz notch and a learned channel adapter; CBraMod used 200 Hz signals filtered at 4--32 Hz with a 60 Hz notch.

\paragraph{Evaluation Protocols and Data Splits.}
For leave-one-subject-out (LOSO) evaluation, one subject was held out for testing, while the remaining subjects were used for model development. For within-subject evaluation, data were divided into development and test partitions using a 90\%/10\% split for BNCI2014001 and SEED and a 30\%/70\% split for CHB-MIT.

\paragraph{Computational Resources.}
The experiments were conducted using eight NVIDIA GeForce RTX 4090 GPUs and one NVIDIA GeForce RTX 4070 GPU.

\section{B. SW-ProxyCE with Cross-Dataset References}
\label{app:cross_dataset_reference}

We further evaluated SW-ProxyCE using reference samples collected from a different dataset and acquisition protocol. The victim models and attacked trials were based on BNCI2014001, while the reference set was obtained exclusively from BCI Competition III Dataset IIIa. This setting introduced a distribution shift between reference construction and target evaluation.

Although both datasets contain the same four-class motor imagery task, they differ in subject populations, acquisition systems, channel configurations, and recording protocols. We performed only the necessary temporal and channel alignment while preserving the remaining dataset-specific characteristics.

We selected one subject from Dataset IIIa as the external reference source. After removing officially marked artifact trials, we retained the largest class-balanced subset with 74 trials per class. These samples were used only for reference modeling and remained disjoint from both victim training and target evaluation.

\begin{table}[t]
\centering
\caption{Cross-dataset attack results on BNCI2014001 using external references from BCI Competition III Dataset IIIa (\(K=74\) per class). All values are subject-averaged BACC (\%) over nine LOSO subjects. Lower attacked BACC and larger BACC drop indicate stronger attacks.}
\label{tab:cross_dataset_attack}
\setlength{\tabcolsep}{4pt}
\begin{tabular}{@{}lcccc@{}}
\toprule
Model & Mode & Clean & Attacked \(\downarrow\) & Drop \(\uparrow\)\\
\midrule
LaBraM    & LP & 47.96 & 33.08 & 14.87\\
LaBraM    & FT & 52.28 & 43.79 & 8.49\\
Neuro-GPT & LP & 41.22 & 20.85 & 20.37\\
Neuro-GPT & FT & 54.88 & 37.48 & 17.40\\
CBraMod   & LP & 45.37 & 12.96 & 32.41\\
CBraMod   & FT & 53.65 & 47.01 & 6.64\\
\bottomrule
\end{tabular}
\end{table}

As shown in Table~\ref{tab:cross_dataset_attack}, SW-ProxyCE reduces the BACC across all six victim configurations despite the cross-dataset distribution shift. The reductions range from \(6.64\) to \(32.41\) percentage points, with an average reduction of \(16.70\) points. These results further highlight the adaptability of SW-ProxyCE: it constructs effective task-aware proxy objectives from task-matched EEG collected under a different acquisition protocol. Moreover, references from only a single external subject remain effective across three encoders and both LP and FT victims, indicating the practical flexibility of the proposed attack under heterogeneous acquisition conditions~\cite{liu2025sdda}.

\section{C. Stability across Random Seeds}

We further evaluated the stability of SW-ProxyCE over 10 random seeds on BNCI2014001 under LOSO evaluation. Each seed produces a different random reference-set sampling, while the victim models, perturbation budget, and other attack settings remain fixed. Table~\ref{tab:seed_stability} reports the mean and standard deviation of the BACC drop across repeated runs. This experiment examines whether the strong transfer performance of SW-ProxyCE is robust to the particular labeled examples selected for reference modeling.

\begin{table}[t]
\centering
\small
\setlength{\tabcolsep}{4.5pt}
\begin{tabular}{@{}llccc@{}}
\toprule
Model & Mode
& Gaussian
& TAA
&  SW-ProxyCE\\
\midrule
LaBraM
& LP & 0.58 $\pm$ 0.51 & 3.85 $\pm$ 0.54 & \textbf{23.21 $\pm$ 3.11}\\
& FT & 0.09 $\pm$ 0.29 & 1.74 $\pm$ 0.33 & \textbf{13.16 $\pm$ 1.54}\\
\addlinespace[2pt]
Neuro-GPT
& LP & 0.57 $\pm$ 0.12 & 5.01 $\pm$ 0.60 & \textbf{22.72 $\pm$ 3.51}\\
& FT & 0.29 $\pm$ 0.13 & 4.61 $\pm$ 0.38 & \textbf{14.79 $\pm$ 2.49}\\
\addlinespace[2pt]
CBraMod
& LP & 0.34 $\pm$ 0.24 & 4.52 $\pm$ 0.46 & \textbf{33.18 $\pm$ 3.22}\\
& FT & $-0.08 \pm 0.10$ & 0.29 $\pm$ 0.34 & \textbf{5.07 $\pm$ 1.89}\\
\bottomrule
\end{tabular}
\caption{Attack stability over 10 random seeds on BNCI2014001 under LOSO evaluation. Results are mean BACC drops $\pm$ standard deviations in percentage points. Larger values indicate stronger attacks.}
\label{tab:seed_stability}
\end{table}

SW-ProxyCE achieves the largest mean BACC reduction in all six victim configurations, covering three encoders and both LP and FT adaptation. Averaged across configurations, it yields a drop of \(18.69\) percentage points, compared with \(3.34\) for TAA and \(0.30\) for Gaussian noise. Its mean advantage over the strongest baseline ranges from \(4.78\) to \(28.66\) points. Moreover, even after accounting for one standard deviation, the lower bound of SW-ProxyCE remains above the corresponding upper bounds of both baselines in every configuration. These results show that the effectiveness of SW-ProxyCE is consistently preserved across different reference samplings rather than being driven by a favorable random selection. Although the attack strength varies moderately with the sampled references, the substantial and persistent margin over TAA and Gaussian noise demonstrates the stable superiority of the proposed task-aware proxy.

\section{D. Sensitivity to the Shrinkage Coefficient}
\label{app:rho_sensitivity}

We examined the sensitivity of SW-ProxyCE to the shrinkage coefficient \(\rho\) on BNCI2014001 under the LOSO protocol. We varied \(\rho\) from \(0\) to \(1\) while fixing \(K=20\), \(\epsilon=0.1\), and all other attack settings. Recall that \(\rho=0\) corresponds to ordinary whitening using the empirical directional variances, whereas increasing \(\rho\) shrinks these variances toward their mean. At \(\rho=1\), all observed within-class directions receive the same variance estimate, removing directional variance differences from the whitening transform.

\begin{figure}[t]
\centering
\includegraphics[width=\columnwidth]{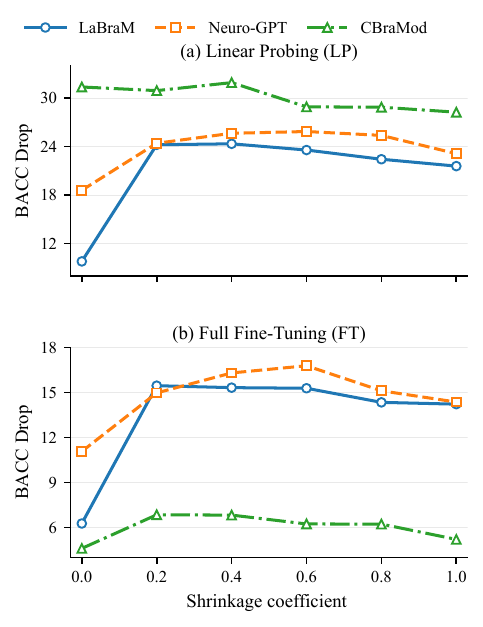}
\caption{Sensitivity of SW-ProxyCE to the shrinkage coefficient \(\rho\) on BNCI2014001 under LOSO evaluation. Panels (a) and (b) show BACC drops against LP and FT victims, respectively. All experiments use \(K=20\) references per class and an \(\ell_\infty\) budget of \(\epsilon=0.1\). Larger BACC drops indicate stronger attacks.}
\label{fig:rho_sensitivity}
\end{figure}

As shown in Figure~\ref{fig:rho_sensitivity}, SW-ProxyCE generally performs best with moderate shrinkage. For LP victims, LaBraM and CBraMod achieve their largest BACC drops at \(\rho=0.4\), while Neuro-GPT reaches its maximum at \(\rho=0.6\). A similar pattern is observed for FT victims: the strongest results occur within the range \(\rho\in[0.2,0.6]\), and the performance differences among these moderate values are relatively small. In contrast, \(\rho=0\) produces substantially weaker attacks for LaBraM and Neuro-GPT, indicating that ordinary whitening can be sensitive to empirical variance estimates obtained from limited references.

Averaged across the three encoders and both adaptation strategies, the BACC drop increases from \(13.62\) percentage points at \(\rho=0\) to \(20.06\) points at \(\rho=0.4\). Further increasing the coefficient gradually reduces the average drop to \(17.80\) points at \(\rho=1\). These results suggest that neither unregularized whitening nor complete removal of directional variance differences is optimal. Moderate shrinkage provides a favorable balance: it preserves informative anisotropic structure while limiting excessive inverse scaling of directions whose variances may be inaccurately estimated from the small reference set. The default choice of \(\rho=0.4\) achieves the strongest average performance and remains close to the optimum in every evaluated configuration, supporting its use as a fixed value throughout the main experiments.

\section{E. Additional Results on ACC and ASR}
\label{sec:appendix_acc_asr}

Table~\ref{tab:appendix_acc_asr} complements the main BACC evaluation with standard classification accuracy (ACC) and attack success rate (ASR). We reported the same datasets, public encoders, downstream adaptation strategies, and evaluation protocols as in the main experiments. BACC remains our primary performance metric since it assigns equal importance to each class, whereas ACC reflects the overall proportion of correct predictions. ASR provides a further attack-oriented measure by quantifying how often an adversarial perturbation overturns a prediction that was initially correct.

\paragraph{Metric definitions.}
Let subject \(s\) contain \(N_s\) test trials, and let \(F_{\mathrm{vic}}\) denote the corresponding private downstream model. The clean ACC for subject \(s\) is defined as
\begin{equation}
\mathrm{ACC}^{\mathrm{clean}}_s
=
\frac{1}{N_s}
\sum_{i=1}^{N_s}
\mathbb{I}\!\left[
F_{\mathrm{vic}}(x_i)=y_i
\right],
\label{eq:appendix_clean_acc}
\end{equation}
where \(x_i\) and \(y_i\) denote the \(i\)-th test trial and its ground-truth label, respectively, and \(\mathbb{I}[\cdot]\) is the indicator function. Given the corresponding adversarial trial \(x_i^{\mathrm{adv}}\), attacked ACC is computed as
\begin{equation}
\mathrm{ACC}^{\mathrm{adv}}_s
=
\frac{1}{N_s}
\sum_{i=1}^{N_s}
\mathbb{I}\!\left[
F_{\mathrm{vic}}(x_i^{\mathrm{adv}})=y_i
\right].
\label{eq:appendix_attacked_acc}
\end{equation}
Both metrics evaluate all test trials. Clean ACC measures the original predictive performance of the private victim, while attacked ACC measures its remaining overall accuracy after perturbation.

ASR instead considers only test trials that are correctly classified before attack:
\begin{equation}
\mathrm{ASR}_s
=
\frac{
\sum_{i=1}^{N_s}
\mathbb{I}\!\left[
F_{\mathrm{vic}}(x_i)=y_i
\land
F_{\mathrm{vic}}(x_i^{\mathrm{adv}})\neq y_i
\right]
}{
\sum_{i=1}^{N_s}
\mathbb{I}\!\left[
F_{\mathrm{vic}}(x_i)=y_i
\right]
}.
\label{eq:appendix_asr}
\end{equation}

The numerator counts clean-correct predictions that are changed into incorrect predictions after attack, while the denominator contains all clean-correct predictions. ASR therefore directly measures the success of the attack on predictions that the victim originally handled correctly. It is not equivalent to the reduction in ACC: an attack may simultaneously change some clean-correct predictions into errors and some clean-incorrect predictions into correct predictions. The latter affects attacked ACC but is not counted by ASR.

Following the main evaluation, all metrics are first computed independently for each subject and then averaged with equal subject weight. For a generic subject-level metric \(M_s\), the reported result is
\begin{equation}
\overline{M}
=
\frac{1}{S}
\sum_{s=1}^{S}M_s,
\label{eq:appendix_subject_average}
\end{equation}
where \(S\) is the number of subjects. This aggregation prevents subjects with more test trials from dominating the reported results.

\paragraph{Results on classification accuracy.}
SW-ProxyCE obtains the lowest attacked ACC in all 36 dataset--encoder--adaptation--protocol configurations. Averaged across these configurations, the clean ACC is \(65.63\%\). Gaussian noise and TAA retain attacked ACC values of \(63.66\%\) and \(58.84\%\), corresponding to average reductions of only \(1.97\) and \(6.79\) percentage points, respectively. In contrast, SW-ProxyCE reduces the average attacked ACC to \(41.41\%\), a decrease of \(24.22\) percentage points from clean performance. The ACC results therefore agree with the main BACC results and show that the observed degradation is not specific to class-balanced evaluation.

The effect is particularly pronounced against LP victims. Across all datasets, encoders, and protocols, SW-ProxyCE yields an average attacked ACC of \(29.69\%\) against LP models, compared with \(53.12\%\) against FT models. This difference indicates that full fine-tuning can reduce transferability by modifying a larger portion of the pretrained representation. Nevertheless, SW-ProxyCE remains the strongest evaluated attack in every FT configuration, showing that full fine-tuning does not eliminate the vulnerability inherited from the public encoder.

\paragraph{Results on attack success rate.}
The ASR results provide a consistent attack-centered view. Averaged across all 36 configurations, Gaussian noise, TAA, and SW-ProxyCE achieve ASRs of \(7.55\%\), \(22.32\%\), and \(43.68\%\), respectively. SW-ProxyCE obtains the highest ASR in every evaluated configuration, demonstrating that its reduction in attacked accuracy primarily reflects successful disruption of predictions that were initially correct rather than incidental changes among already misclassified samples.

The LP--FT comparison follows the same pattern observed for attacked ACC. SW-ProxyCE reaches an average ASR of \(59.85\%\) against LP victims and \(27.51\%\) against FT victims. Thus, full fine-tuning provides greater resistance to transferred perturbations, but a substantial fraction of initially correct predictions can still be overturned without accessing or querying the private model. The consistent trends under both LOSO and within-subject evaluation further show that the attack is effective across cross-subject transfer and subject-specific adaptation settings.

Overall, BACC, ACC, and ASR characterize complementary aspects of the same vulnerability. BACC measures class-balanced post-attack performance, ACC measures overall post-attack correctness, and ASR isolates the failure of predictions that were correct before perturbation. Their consistent results provide converging evidence that adversarial examples generated through the public encoder and a small labeled reference set can transfer effectively to private LP and FT downstream models.
%\clearpage
\begin{table*}[p]
\centering
\begin{tabular*}{\textwidth}{@{}ll@{\extracolsep{\fill}}cccc@{}}
\toprule
\multirow{2}{*}{Dataset}
& \multirow{2}{*}{Model}
& \multirow{2}{*}{Clean ACC}
& \multicolumn{3}{c}{Attacked ACC \(\downarrow\) / ASR \(\uparrow\)}\\
\cmidrule(lr){4-6}
& & & GN & TAA & Ours\\
\midrule
\multicolumn{6}{c}{\textbf{LOSO}}\\
\addlinespace[2pt]
\multicolumn{6}{@{}l@{}}{\textbf{Linear Probing (LP)}}\\
\addlinespace[2pt]
\multirow{3}{*}{BNCI2014001}
& LaBraM
& 47.96
& 47.55 / 14.76
& 43.92 / 30.95
& \textbf{23.61 / 58.20}\\
& Neuro-GPT
& 41.22
& 40.91 / 5.41
& 37.23 / 35.13
& \textbf{15.57 / 65.17}\\
& CBraMod
& 45.37
& 45.33 / 3.59
& 41.44 / 31.00
& \textbf{13.45 / 72.57}\\
\cmidrule{1-6}
\multirow{3}{*}{SEED}
& LaBraM
& 54.12
& 46.29 / 43.30
& 35.74 / 66.61
& \textbf{33.10 / 71.04}\\
& Neuro-GPT
& 46.06
& 46.01 / 1.81
& 44.23 / 20.33
& \textbf{23.85 / 48.97}\\
& CBraMod
& 51.97
& 51.93 / 4.02
& 50.68 / 17.30
& \textbf{36.19 / 34.54}\\
\cmidrule{1-6}
\multirow{3}{*}{CHB-MIT}
& LaBraM
& 86.29
& 79.76 / 12.64
& 60.55 / 36.98
& \textbf{31.36 / 67.22}\\
& Neuro-GPT
& 87.77
& 87.73 / 0.29
& 81.54 / 11.98
& \textbf{11.78 / 87.04}\\
& CBraMod
& 66.06
& 68.08 / 1.99
& 69.21 / 10.58
& \textbf{49.76 / 31.00}\\
\midrule
\multicolumn{6}{@{}l@{}}{\textbf{Full Fine-Tuning (FT)}}\\
\addlinespace[2pt]
\multirow{3}{*}{BNCI2014001}
& LaBraM
& 52.28
& 51.99 / 6.92
& 51.22 / 17.28
& \textbf{36.96 / 34.92}\\
& Neuro-GPT
& 54.88
& 54.46 / 3.54
& 50.15 / 21.88
& \textbf{38.58 / 34.21}\\
& CBraMod
& 53.65
& 53.74 / 2.08
& 53.74 / 5.96
& \textbf{46.82 / 15.08}\\
\cmidrule{1-6}
\multirow{3}{*}{SEED}
& LaBraM
& 55.61
& 51.38 / 28.09
& 50.75 / 40.36
& \textbf{48.21 / 41.29}\\
& Neuro-GPT
& 53.66
& 53.60 / 2.78
& 53.42 / 7.96
& \textbf{48.61 / 13.00}\\
& CBraMod
& 53.03
& 52.62 / 10.24
& 48.99 / 32.12
& \textbf{45.14 / 35.29}\\
\cmidrule{1-6}
\multirow{3}{*}{CHB-MIT}
& LaBraM
& 78.69
& 75.98 / 7.47
& 71.52 / 16.60
& \textbf{59.80 / 28.63}\\
& Neuro-GPT
& 80.54
& 80.45 / 0.53
& 81.46 / 2.87
& \textbf{76.54 / 6.81}\\
& CBraMod
& 68.75
& 68.44 / 2.25
& 69.31 / 2.86
& \textbf{64.11 / 8.54}\\
\midrule
\multicolumn{6}{c}{\textbf{Within-Subject}}\\
\addlinespace[2pt]
\multicolumn{6}{@{}l@{}}{\textbf{Linear Probing (LP)}}\\
\addlinespace[2pt]
\multirow{3}{*}{BNCI2014001}
& LaBraM
& 54.37
& 54.37 / 12.62
& 46.03 / 35.89
& \textbf{2.18 / 96.30}\\
& Neuro-GPT
& 51.59
& 50.79 / 4.58
& 36.90 / 43.48
& \textbf{11.11 / 79.90}\\
& CBraMod
& 50.20
& 50.00 / 3.13
& 48.41 / 26.00
& \textbf{1.39 / 97.20}\\
\cmidrule{1-6}
\multirow{3}{*}{SEED}
& LaBraM
& 78.40
& 61.22 / 31.10
& 40.52 / 58.43
& \textbf{29.51 / 69.50}\\
& Neuro-GPT
& 50.37
& 50.35 / 1.58
& 47.26 / 22.53
& \textbf{27.58 / 49.59}\\
& CBraMod
& 74.90
& 74.60 / 2.47
& 69.83 / 14.96
& \textbf{24.92 / 67.18}\\
\cmidrule{1-6}
\multirow{3}{*}{CHB-MIT}
& LaBraM
& 94.82
& 79.86 / 17.31
& 70.75 / 26.84
& \textbf{63.26 / 34.18}\\
& Neuro-GPT
& 82.20
& 82.29 / 0.48
& 80.59 / 7.26
& \textbf{52.37 / 37.39}\\
& CBraMod
& 92.49
& 92.53 / 0.38
& 91.34 / 2.94
& \textbf{83.41 / 10.23}\\
\midrule
\multicolumn{6}{@{}l@{}}{\textbf{Full Fine-Tuning (FT)}}\\
\addlinespace[2pt]
\multirow{3}{*}{BNCI2014001}
& LaBraM
& 49.80
& 49.01 / 14.11
& 42.46 / 37.39
& \textbf{4.56 / 91.29}\\
& Neuro-GPT
& 50.20
& 50.00 / 2.29
& 41.27 / 30.98
& \textbf{25.99 / 51.95}\\
& CBraMod
& 56.35
& 56.35 / 1.24
& 55.16 / 8.03
& \textbf{49.01 / 14.34}\\
\cmidrule{1-6}
\multirow{3}{*}{SEED}
& LaBraM
& 86.12
& 71.51 / 21.47
& 46.82 / 50.76
& \textbf{45.29 / 52.37}\\
& Neuro-GPT
& 73.86
& 73.77 / 1.67
& 71.28 / 10.26
& \textbf{64.42 / 17.46}\\
& CBraMod
& 87.09
& 85.68 / 3.41
& 81.70 / 9.96
& \textbf{72.57 / 18.84}\\
\cmidrule{1-6}
\multirow{3}{*}{CHB-MIT}
& LaBraM
& 74.87
& 75.53 / 1.82
& 74.89 / 3.89
& \textbf{73.10 / 5.86}\\
& Neuro-GPT
& 90.50
& 90.56 / 0.08
& 91.17 / 3.36
& \textbf{74.25 / 19.90}\\
& CBraMod
& 86.67
& 87.23 / 0.29
& 86.78 / 1.72
& \textbf{82.26 / 5.37}\\
\bottomrule
\end{tabular*}
\caption{Additional transfer attack results under LOSO and within-subject evaluation. Clean results are reported as subject-averaged ACC (\%). Each attacked entry reports subject-averaged attacked ACC and ASR, respectively, in the form ACC/ASR (\%). ASR is computed over samples correctly classified before attack. Lower attacked ACC and higher ASR indicate stronger attacks. GN denotes Gaussian noise, and Ours denotes SW-ProxyCE. The strongest attack result in each setting is shown in bold.}
\label{tab:appendix_acc_asr}
\end{table*}
%\clearpage

\section{F. Effect of Perturbation Budget}
\label{sec:epsilon_ablation}

We examined the sensitivity of SW-ProxyCE to the perturbation budget by varying \(\epsilon\in\{0.05,0.10,0.20\}\). The reference-set size is fixed at \(K=20\) samples per class, and all remaining attack settings are kept unchanged. We evaluated all three datasets, three public encoders, two downstream adaptation strategies, and both LOSO and within-subject protocols, resulting in 36 configurations. Table~\ref{tab:epsilon_ablation} reports subject-averaged ACC and BACC before and after attack. The same perturbation budget is applied to every dataset, encoder, and private victim without configuration-specific tuning.

Attack effectiveness generally increases with the perturbation budget. Averaged over all 36 configurations, attacked ACC decreases from \(49.37\%\) at \(\epsilon=0.05\) to \(41.41\%\) at \(\epsilon=0.10\) and \(35.89\%\) at \(\epsilon=0.20\). The corresponding attacked BACC values decrease from \(44.82\%\) to \(37.50\%\) and \(32.15\%\), respectively. For comparison, the average clean ACC and BACC are \(65.63\%\) and \(60.40\%\). Thus, even the smallest evaluated budget causes average reductions of \(16.26\) percentage points in ACC and \(15.58\) percentage points in BACC, showing that SW-ProxyCE remains effective under a relatively restricted perturbation budget.

The trend is consistent across the two evaluation protocols. Under LOSO, the mean attacked BACC decreases from \(41.97\%\) at \(\epsilon=0.05\) to \(35.88\%\) at \(\epsilon=0.10\) and \(30.61\%\) at \(\epsilon=0.20\). Under within-subject evaluation, the corresponding values are \(47.68\%\), \(39.11\%\), and \(33.70\%\). Attacked ACC follows the same overall pattern, decreasing from \(46.33\%\) to \(39.08\%\) and \(33.36\%\) under LOSO, and from \(52.42\%\) to \(43.73\%\) and \(38.42\%\) under within-subject evaluation. These results show that the budget-dependent increase in attack strength is not confined to either cross-subject transfer or subject-specific adaptation.

The effect is particularly pronounced against LP victims. Their mean attacked BACC decreases from \(34.70\%\) at \(\epsilon=0.05\) to \(25.53\%\) at \(\epsilon=0.10\) and \(21.60\%\) at \(\epsilon=0.20\). For FT victims, the corresponding values are \(54.94\%\), \(49.47\%\), and \(42.71\%\). A similar gap appears in ACC: increasing \(\epsilon\) from \(0.05\) to \(0.20\) reduces the mean attacked ACC from \(40.21\%\) to \(25.21\%\) for LP victims and from \(58.53\%\) to \(46.57\%\) for FT victims. This difference is consistent with full fine-tuning altering a larger portion of the pretrained representation and thereby weakening transfer from the public encoder. Nevertheless, the persistent degradation at all three budgets shows that full fine-tuning mitigates, but does not remove, the transfer vulnerability.

At the individual-configuration level, BACC decreases monotonically with increasing \(\epsilon\) in 35 of the 36 configurations, while ACC does so in 34 configurations. The main exception is the BNCI2014001 within-subject CBraMod-LP setting, where \(\epsilon=0.10\) produces a lower attacked ACC and BACC than \(\epsilon=0.20\). A small ACC reversal also occurs for CHB-MIT CBraMod-LP under LOSO between \(\epsilon=0.10\) and \(0.20\), although its BACC continues to decrease. Such isolated non-monotonic behavior is possible because perturbations are optimized against the public-encoder proxy rather than the private victim itself; a larger displacement in the proxy space does not necessarily induce a strictly larger change in every inaccessible victim decision function. The dominant trend across configurations nevertheless shows that increasing the available perturbation budget generally improves transfer effectiveness.

We used \(\epsilon=0.10\) for the main experiments as a fixed intermediate budget. This choice avoids selecting a separate budget for each dataset or victim while providing a nontrivial attack strength between the more constrained \(\epsilon=0.05\) setting and the stronger \(\epsilon=0.20\) setting. The results across all three budgets further demonstrate that the observed transfer risk is not tied to a single perturbation magnitude.

%\clearpage
\begin{table*}[p]
\centering
\begin{tabular*}{\textwidth}{@{}ll@{\extracolsep{\fill}}cccc@{}}
\toprule
\multirow{2}{*}{Dataset}
& \multirow{2}{*}{Model}
& \multirow{2}{*}{Clean ACC / BACC}
& \multicolumn{3}{c}{Attacked ACC / BACC \(\downarrow\)}\\
\cmidrule(lr){4-6}
& & & \(\epsilon=0.05\) & \(\epsilon=0.10\) & \(\epsilon=0.20\)\\
\midrule
\multicolumn{6}{c}{\textbf{LOSO}}\\
\addlinespace[2pt]
\multicolumn{6}{@{}l@{}}{\textbf{Linear Probing (LP)}}\\
\addlinespace[2pt]
\multirow{3}{*}{BNCI2014001}
& LaBraM & 47.96 / 47.96 & 29.13 / 29.13 & 23.61 / 23.61 & 20.06 / 20.06\\
& Neuro-GPT & 41.22 / 41.22 & 26.02 / 26.02 & 15.57 / 15.57 & 9.76 / 9.76\\
& CBraMod & 45.37 / 45.37 & 22.61 / 22.61 & 13.45 / 13.45 & 6.21 / 6.21\\
\cmidrule{1-6}
\multirow{3}{*}{SEED}
& LaBraM & 54.12 / 53.87 & 35.12 / 35.29 & 33.10 / 33.37 & 32.40 / 32.69\\
& Neuro-GPT & 46.06 / 45.96 & 33.36 / 33.29 & 23.85 / 23.75 & 15.43 / 15.21\\
& CBraMod & 51.97 / 51.71 & 41.63 / 41.38 & 36.19 / 35.99 & 30.13 / 30.10\\
\cmidrule{1-6}
\multirow{3}{*}{CHB-MIT}
& LaBraM & 86.29 / 55.64 & 43.78 / 34.95 & 31.36 / 31.15 & 23.19 / 26.79\\
& Neuro-GPT & 87.77 / 65.36 & 45.70 / 29.08 & 11.78 / 7.21 & 2.17 / 1.19\\
& CBraMod & 66.06 / 54.00 & 54.33 / 39.11 & 49.76 / 33.78 & 50.14 / 32.71\\
\midrule
\multicolumn{6}{@{}l@{}}{\textbf{Full Fine-Tuning (FT)}}\\
\addlinespace[2pt]
\multirow{3}{*}{BNCI2014001}
& LaBraM & 52.28 / 52.28 & 42.53 / 42.53 & 36.96 / 36.96 & 30.61 / 30.61\\
& Neuro-GPT & 54.88 / 54.88 & 47.09 / 47.09 & 38.58 / 38.58 & 26.47 / 26.47\\
& CBraMod & 53.65 / 53.65 & 50.15 / 50.15 & 46.82 / 46.82 & 40.72 / 40.72\\
\cmidrule{1-6}
\multirow{3}{*}{SEED}
& LaBraM & 55.61 / 55.37 & 49.61 / 49.24 & 48.21 / 47.67 & 44.68 / 44.02\\
& Neuro-GPT & 53.66 / 53.40 & 51.08 / 50.80 & 48.61 / 48.31 & 45.90 / 45.57\\
& CBraMod & 53.03 / 52.71 & 50.34 / 50.08 & 45.14 / 44.98 & 38.37 / 38.30\\
\cmidrule{1-6}
\multirow{3}{*}{CHB-MIT}
& LaBraM & 78.69 / 58.95 & 67.26 / 52.97 & 59.80 / 48.69 & 48.11 / 43.55\\
& Neuro-GPT & 80.54 / 68.46 & 78.44 / 65.93 & 76.54 / 62.86 & 74.59 / 58.04\\
& CBraMod & 68.75 / 58.80 & 65.69 / 55.78 & 64.11 / 53.16 & 61.54 / 48.96\\
\midrule
\multicolumn{6}{c}{\textbf{Within-Subject}}\\
\addlinespace[2pt]
\multicolumn{6}{@{}l@{}}{\textbf{Linear Probing (LP)}}\\
\addlinespace[2pt]
\multirow{3}{*}{BNCI2014001}
& LaBraM & 54.37 / 54.37 & 4.37 / 4.37 & 2.18 / 2.18 & 0.79 / 0.79\\
& Neuro-GPT & 51.59 / 51.59 & 24.21 / 24.21 & 11.11 / 11.11 & 10.52 / 10.52\\
& CBraMod & 50.20 / 50.20 & 25.79 / 25.79 & 1.39 / 1.39 & 13.89 / 13.89\\
\cmidrule{1-6}
\multirow{3}{*}{SEED}
& LaBraM & 78.40 / 78.25 & 33.27 / 33.18 & 29.51 / 29.47 & 28.92 / 28.89\\
& Neuro-GPT & 50.37 / 50.12 & 37.30 / 37.11 & 27.58 / 27.41 & 18.32 / 18.11\\
& CBraMod & 74.90 / 74.73 & 40.76 / 40.55 & 24.92 / 24.88 & 15.18 / 15.35\\
\cmidrule{1-6}
\multirow{3}{*}{CHB-MIT}
& LaBraM & 94.82 / 70.11 & 70.69 / 53.29 & 63.26 / 50.45 & 60.97 / 49.25\\
& Neuro-GPT & 82.20 / 70.46 & 67.65 / 54.41 & 52.37 / 39.79 & 38.96 / 27.47\\
& CBraMod & 92.49 / 70.52 & 88.06 / 60.84 & 83.41 / 54.93 & 76.75 / 49.81\\
\midrule
\multicolumn{6}{@{}l@{}}{\textbf{Full Fine-Tuning (FT)}}\\
\addlinespace[2pt]
\multirow{3}{*}{BNCI2014001}
& LaBraM & 49.80 / 49.80 & 11.11 / 11.11 & 4.56 / 4.56 & 1.98 / 1.98\\
& Neuro-GPT & 50.20 / 50.20 & 38.10 / 38.10 & 25.99 / 25.99 & 20.44 / 20.44\\
& CBraMod & 56.35 / 56.35 & 52.18 / 52.18 & 49.01 / 49.01 & 45.63 / 45.63\\
\cmidrule{1-6}
\multirow{3}{*}{SEED}
& LaBraM & 86.12 / 86.04 & 56.03 / 55.61 & 45.29 / 44.93 & 37.56 / 37.30\\
& Neuro-GPT & 73.86 / 73.65 & 69.70 / 69.43 & 64.42 / 64.10 & 54.85 / 54.49\\
& CBraMod & 87.09 / 87.00 & 80.54 / 80.44 & 72.57 / 72.46 & 57.90 / 57.80\\
\cmidrule{1-6}
\multirow{3}{*}{CHB-MIT}
& LaBraM & 74.87 / 64.02 & 75.60 / 62.65 & 73.10 / 60.62 & 72.16 / 56.39\\
& Neuro-GPT & 90.50 / 85.51 & 83.87 / 75.70 & 74.25 / 63.40 & 58.55 / 45.51\\
& CBraMod & 86.67 / 81.80 & 84.30 / 79.21 & 82.26 / 77.37 & 78.15 / 72.94\\
\bottomrule
\end{tabular*}
\caption{Effect of the perturbation budget under LOSO and within-subject evaluation. Each entry reports subject-averaged ACC/BACC (\%) with \(K=20\) references per class. All attack settings other than the \(\ell_\infty\) budget \(\epsilon\) are held fixed. Lower attacked ACC and BACC indicate stronger attacks.}
\label{tab:epsilon_ablation}
\end{table*}

\section{G. SW-ProxyCE under Different Attack Optimizers}
\label{app:alternative_optimizers}

The main experiments instantiate SW-ProxyCE using the Fast Gradient
Sign Method (FGSM). To determine whether its transferability depends
on a particular gradient update rule, we further combined the same
proxy objective with momentum iterative (MI) and projected gradient
descent (PGD) optimization \citep{MI,PGD}. Across all variants, the public encoder,
labeled reference set, shrinkage-whitened proxy geometry, perturbation
constraint, and private downstream victims remain unchanged. The
variants differ only in how the gradient of the SW-ProxyCE objective
is used to update the adversarial input. None of them requires access
to or queries of the private victim.

\paragraph{Fast Gradient Sign Method.}
FGSM constructs an adversarial example using a single update along the
sign direction of the input gradient. Given a clean EEG trial \(x\)
and its label \(y\), the adversarial input is generated as
\begin{equation}
x^{\mathrm{adv}}
=
x
+
\epsilon
\operatorname{sign}
\left(
\nabla_x
\mathcal{L}_{\mathrm{SW\text{-}ProxyCE}}(x,y)
\right),
\label{eq:appendix_fgsm}
\end{equation}
where \(\epsilon\) denotes the \(\ell_\infty\) perturbation budget.
FGSM applies the full budget in one step and requires only one gradient
evaluation. Its performance in the main experiments shows that the
public encoder and limited labeled references already expose
transferable adversarial directions.

\paragraph{Momentum Iterative Optimization.}
The MI variant performs multiple updates while accumulating gradient
information across iterations. Let \(x^{(0)}=x\) and \(g^{(0)}=0\).
At iteration \(t\), the accumulated gradient is updated as
\begin{equation}
g^{(t+1)}
=
\mu g^{(t)}
+
\frac{
\nabla_{x^{(t)}}
\mathcal{L}_{\mathrm{SW\text{-}ProxyCE}}
\left(x^{(t)},y\right)
}{
\left\|
\nabla_{x^{(t)}}
\mathcal{L}_{\mathrm{SW\text{-}ProxyCE}}
\left(x^{(t)},y\right)
\right\|_1
},
\label{eq:appendix_mi_gradient}
\end{equation}
and the adversarial input is updated by
\begin{equation}
x^{(t+1)}
=
\Pi_{\mathcal{B}_{\epsilon}(x)}
\left[
x^{(t)}
+
\alpha
\operatorname{sign}
\left(g^{(t+1)}\right)
\right],
\label{eq:appendix_mi_update}
\end{equation}
where \(\mu\) is the momentum factor, \(\alpha\) is the step size, and
\(\Pi_{\mathcal{B}_{\epsilon}(x)}\) projects the updated input onto the
\(\ell_\infty\) ball of radius \(\epsilon\) centered at \(x\).
Accumulating normalized gradients encourages a stable optimization
direction across iterations.

\paragraph{Projected Gradient Descent.}
The PGD variant applies iterative sign-gradient updates followed by
projection. Starting from \(x^{(0)}=x\), we first compute
\begin{equation}
g^{(t)}
=
\nabla_{x^{(t)}}
\mathcal{L}_{\mathrm{SW\text{-}ProxyCE}}
\left(x^{(t)},y\right).
\label{eq:appendix_pgd_gradient}
\end{equation}
The adversarial example is then updated as
\begin{equation}
x^{(t+1)}
=
\Pi_{\mathcal{B}_{\epsilon}(x)}
\left[
x^{(t)}
+
\alpha\operatorname{sign}\left(g^{(t)}\right)
\right].
\label{eq:appendix_pgd_update}
\end{equation}
Unlike FGSM, PGD progressively maximizes the proxy objective over
multiple projected updates. It therefore tests whether the proxy
objective remains effective beyond a single local gradient step.

\paragraph{Experimental Settings.}
We evaluated MI and PGD under both LOSO and within-subject protocols on
BNCI2014001, SEED, and CHB-MIT. The evaluation covers LaBraM,
Neuro-GPT, and CBraMod with both linear probing and full fine-tuning.
All attacks use \(K=20\) labeled references per class and an
\(\ell_\infty\) budget of \(\epsilon=0.1\). The reference partitions,
victim models, shrinkage coefficient, proxy construction, and
preprocessing remain fixed across optimizers. We reported
subject-averaged clean and attacked ACC/BACC. Lower attacked values
indicate stronger transfer.

\paragraph{Results with Momentum Iterative Optimization.}
Table~\ref{tab:appendix_mi_acc_bacc} presents the MI results.
SW-ProxyCE-MI achieves the lowest attacked ACC and BACC in all 36
configurations. Under within-subject evaluation, it reduces the BACC
of LaBraM-LP on BNCI2014001 from \(54.37\%\) to \(0.00\%\), and those
of LaBraM-LP and CBraMod-LP on SEED from \(78.25\%\) and \(74.73\%\)
to \(5.28\%\) and \(15.42\%\), respectively.

The attack also transfers strongly on CHB-MIT. Under LOSO, the BACC of
LaBraM-LP decreases from \(55.64\%\) to \(13.39\%\), while that of
Neuro-GPT-LP decreases from \(65.36\%\) to \(6.51\%\). Under
within-subject evaluation, LaBraM-LP is reduced from \(70.11\%\) to
\(22.71\%\). Although FT victims are generally more resistant, clear
degradation remains across all three datasets.

\paragraph{Results with Projected Gradient Descent.}
Table~\ref{tab:appendix_pgd_acc_bacc} reports the PGD results.
SW-ProxyCE-PGD likewise obtains the lowest attacked ACC and BACC in
all 36 configurations. On BNCI2014001 within subjects, the attacked
BACC reaches \(0.00\%\), \(9.92\%\), and \(12.90\%\) for LaBraM,
Neuro-GPT, and CBraMod LP victims, respectively. On SEED, the
corresponding BACC values for LaBraM-LP and CBraMod-LP are \(6.17\%\)
and \(13.92\%\).

The CHB-MIT results show the same behavior. Under LOSO, PGD reduces
Neuro-GPT-LP from \(65.36\%\) to \(5.16\%\) BACC and LaBraM-LP from
\(55.64\%\) to \(15.50\%\). Under within-subject evaluation, LaBraM-LP
and Neuro-GPT-LP are reduced to \(25.05\%\) and \(37.80\%\) BACC.
These results confirm that the proxy objective remains effective under
repeated projected optimization.

\paragraph{Comparison across Attack Optimizers.}
Across the 36 matched SW-ProxyCE configurations, MI obtains lower
attacked BACC than PGD in 22 cases, PGD performs better in 12 cases,
and 2 cases are tied. Their mean attacked BACC values are \(33.73\%\)
and \(34.35\%\), respectively. Thus, MI has a modest average advantage,
but neither iterative optimizer uniformly dominates. More importantly,
both variants consistently outperform GN and their optimizer-matched
TAA baselines, indicating that transferability primarily arises from
the shrinkage-whitened proxy objective rather than a specific gradient
update rule.

Together with the FGSM results, these experiments show that SW-ProxyCE
retains its effectiveness under single-step, momentum-based, and
projected iterative optimization.

\begin{table*}[p]
\centering
\begin{tabular*}{\textwidth}{@{}ll@{\extracolsep{\fill}}cccc@{}}
\toprule
\multirow{2}{*}{Dataset}
& \multirow{2}{*}{Model}
& \multirow{2}{*}{Clean ACC / BACC}
& \multicolumn{3}{c}{Attacked ACC \(\downarrow\) / BACC \(\downarrow\)}\\
\cmidrule(lr){4-6}
& & & GN & TAA & Ours\\
\midrule
\multicolumn{6}{c}{\textbf{LOSO}}\\
\addlinespace[2pt]
\multicolumn{6}{@{}l@{}}{\textbf{Linear Probing (LP)}}\\
\addlinespace[2pt]

\multirow{3}{*}{BNCI2014001}
& LaBraM
& 47.96 / 47.96
& 47.55 / 47.55
& 43.29 / 43.29
& \textbf{18.27 / 18.27}\\
& Neuro-GPT
& 41.22 / 41.22
& 40.91 / 40.91
& 35.94 / 35.94
& \textbf{14.26 / 14.26}\\
& CBraMod
& 45.37 / 45.37
& 45.33 / 45.33
& 39.64 / 39.64
& \textbf{10.47 / 10.47}\\

\cmidrule{1-6}
\multirow{3}{*}{SEED}
& LaBraM
& 54.12 / 53.87
& 46.29 / 46.11
& 34.16 / 34.35
& \textbf{27.59 / 27.87}\\
& Neuro-GPT
& 46.06 / 45.96
& 46.01 / 45.90
& 44.05 / 43.83
& \textbf{23.63 / 23.53}\\
& CBraMod
& 51.97 / 51.70
& 51.93 / 51.66
& 50.44 / 50.20
& \textbf{33.85 / 33.65}\\

\cmidrule{1-6}
\multirow{3}{*}{CHB-MIT}
& LaBraM
& 86.29 / 55.64
& 79.76 / 56.78
& 78.63 / 53.22
& \textbf{16.34 / 13.39}\\
& Neuro-GPT
& 87.77 / 65.36
& 87.73 / 65.49
& 78.41 / 60.87
& \textbf{10.71 / 6.51}\\
& CBraMod
& 66.06 / 54.00
& 68.08 / 53.79
& 68.96 / 52.29
& \textbf{43.17 / 28.63}\\

\midrule
\multicolumn{6}{@{}l@{}}{\textbf{Full Fine-Tuning (FT)}}\\
\addlinespace[2pt]

\multirow{3}{*}{BNCI2014001}
& LaBraM
& 52.28 / 52.28
& 51.99 / 51.99
& 51.08 / 51.08
& \textbf{36.30 / 36.30}\\
& Neuro-GPT
& 54.88 / 54.88
& 54.46 / 54.46
& 49.73 / 49.73
& \textbf{38.95 / 38.95}\\
& CBraMod
& 53.65 / 53.65
& 53.74 / 53.74
& 53.70 / 53.70
& \textbf{46.33 / 46.33}\\

\cmidrule{1-6}
\multirow{3}{*}{SEED}
& LaBraM
& 55.61 / 55.37
& 51.38 / 51.07
& 48.18 / 47.72
& \textbf{46.37 / 45.92}\\
& Neuro-GPT
& 53.66 / 53.40
& 53.60 / 53.33
& 53.45 / 53.19
& \textbf{48.57 / 48.27}\\
& CBraMod
& 53.03 / 52.71
& 52.62 / 52.31
& 49.38 / 49.24
& \textbf{44.05 / 43.85}\\

\cmidrule{1-6}
\multirow{3}{*}{CHB-MIT}
& LaBraM
& 78.69 / 58.95
& 75.98 / 59.16
& 73.88 / 58.71
& \textbf{59.35 / 44.33}\\
& Neuro-GPT
& 80.54 / 68.46
& 80.45 / 68.38
& 81.84 / 67.30
& \textbf{76.56 / 63.12}\\
& CBraMod
& 68.75 / 58.80
& 68.44 / 58.64
& 69.16 / 58.52
& \textbf{64.09 / 53.26}\\

\midrule
\multicolumn{6}{c}{\textbf{Within-Subject}}\\
\addlinespace[2pt]
\multicolumn{6}{@{}l@{}}{\textbf{Linear Probing (LP)}}\\
\addlinespace[2pt]

\multirow{3}{*}{BNCI2014001}
& LaBraM
& 54.37 / 54.37
& 54.37 / 54.37
& 45.04 / 45.04
& \textbf{0.00 / 0.00}\\
& Neuro-GPT
& 51.59 / 51.59
& 50.79 / 50.79
& 37.90 / 37.90
& \textbf{9.92 / 9.92}\\
& CBraMod
& 50.20 / 50.20
& 50.00 / 50.00
& 42.66 / 42.66
& \textbf{13.89 / 13.89}\\

\cmidrule{1-6}
\multirow{3}{*}{SEED}
& LaBraM
& 78.40 / 78.25
& 61.22 / 60.96
& 34.80 / 34.78
& \textbf{5.24 / 5.28}\\
& Neuro-GPT
& 50.37 / 50.12
& 50.35 / 50.09
& 47.06 / 46.74
& \textbf{26.93 / 26.75}\\
& CBraMod
& 74.90 / 74.73
& 74.60 / 74.43
& 68.22 / 68.13
& \textbf{15.50 / 15.42}\\

\cmidrule{1-6}
\multirow{3}{*}{CHB-MIT}
& LaBraM
& 94.82 / 70.11
& 79.86 / 62.32
& 65.43 / 52.75
& \textbf{39.91 / 22.71}\\
& Neuro-GPT
& 82.20 / 70.46
& 82.29 / 70.27
& 79.48 / 65.70
& \textbf{51.35 / 39.02}\\
& CBraMod
& 92.49 / 70.52
& 92.53 / 70.37
& 91.30 / 68.59
& \textbf{77.60 / 48.74}\\

\midrule
\multicolumn{6}{@{}l@{}}{\textbf{Full Fine-Tuning (FT)}}\\
\addlinespace[2pt]

\multirow{3}{*}{BNCI2014001}
& LaBraM
& 49.80 / 49.80
& 49.01 / 49.01
& 41.07 / 41.07
& \textbf{1.59 / 1.59}\\
& Neuro-GPT
& 50.20 / 50.20
& 50.00 / 50.00
& 39.09 / 39.09
& \textbf{26.19 / 26.19}\\
& CBraMod
& 56.35 / 56.35
& 56.35 / 55.95
& 54.76 / 54.76
& \textbf{47.82 / 47.82}\\

\cmidrule{1-6}
\multirow{3}{*}{SEED}
& LaBraM
& 86.12 / 86.04
& 71.51 / 71.23
& 46.97 / 46.68
& \textbf{31.85 / 31.54}\\
& Neuro-GPT
& 73.86 / 73.65
& 73.77 / 73.54
& 71.30 / 71.06
& \textbf{64.38 / 64.06}\\
& CBraMod
& 87.09 / 87.00
& 85.68 / 85.59
& 81.55 / 81.46
& \textbf{68.91 / 68.77}\\

\cmidrule{1-6}
\multirow{3}{*}{CHB-MIT}
& LaBraM
& 74.87 / 64.02
& 75.53 / 63.49
& 74.99 / 60.05
& \textbf{73.60 / 57.99}\\
& Neuro-GPT
& 90.50 / 85.51
& 90.56 / 85.61
& 89.82 / 83.71
& \textbf{72.84 / 62.67}\\
& CBraMod
& 86.67 / 81.80
& 87.23 / 81.73
& 86.58 / 81.11
& \textbf{80.37 / 74.96}\\

\bottomrule
\end{tabular*}
\caption{Transfer attack results using momentum iterative optimization under LOSO and within-subject evaluation. Each entry reports subject-averaged ACC/BACC (\%). TAA and SW-ProxyCE use the MI-style optimizer. Lower attacked values indicate stronger attacks, and the strongest attacked result in each setting is shown in bold. GN denotes Gaussian noise, and Ours denotes SW-ProxyCE.}
\label{tab:appendix_mi_acc_bacc}
\end{table*}

\begin{table*}[p]
\centering
\begin{tabular*}{\textwidth}{@{}ll@{\extracolsep{\fill}}cccc@{}}
\toprule
\multirow{2}{*}{Dataset}
& \multirow{2}{*}{Model}
& \multirow{2}{*}{Clean ACC / BACC}
& \multicolumn{3}{c}{Attacked ACC \(\downarrow\) / BACC \(\downarrow\)}\\
\cmidrule(lr){4-6}
& & & GN & TAA & Ours\\
\midrule
\multicolumn{6}{c}{\textbf{LOSO}}\\
\addlinespace[2pt]
\multicolumn{6}{@{}l@{}}{\textbf{Linear Probing (LP)}}\\
\addlinespace[2pt]

\multirow{3}{*}{BNCI2014001}
& LaBraM
& 47.96 / 47.96
& 47.55 / 47.55
& 44.70 / 44.70
& \textbf{22.78 / 22.78}\\
& Neuro-GPT
& 41.22 / 41.22
& 40.91 / 40.91
& 35.46 / 35.46
& \textbf{15.35 / 15.35}\\
& CBraMod
& 45.37 / 45.37
& 45.33 / 45.33
& 38.97 / 38.97
& \textbf{10.92 / 10.92}\\

\cmidrule{1-6}
\multirow{3}{*}{SEED}
& LaBraM
& 54.12 / 53.87
& 46.29 / 46.11
& 34.73 / 34.86
& \textbf{25.19 / 25.43}\\
& Neuro-GPT
& 46.06 / 45.96
& 46.01 / 45.90
& 43.85 / 43.62
& \textbf{23.88 / 23.76}\\
& CBraMod
& 51.97 / 51.70
& 51.93 / 51.66
& 50.29 / 50.03
& \textbf{34.22 / 34.04}\\

\cmidrule{1-6}
\multirow{3}{*}{CHB-MIT}
& LaBraM
& 86.29 / 55.64
& 79.76 / 56.78
& 82.06 / 50.63
& \textbf{21.81 / 15.50}\\
& Neuro-GPT
& 87.77 / 65.36
& 87.73 / 65.49
& 77.20 / 59.87
& \textbf{8.32 / 5.16}\\
& CBraMod
& 66.06 / 54.00
& 68.08 / 53.79
& 68.48 / 52.18
& \textbf{42.19 / 28.27}\\

\midrule
\multicolumn{6}{@{}l@{}}{\textbf{Full Fine-Tuning (FT)}}\\
\addlinespace[2pt]

\multirow{3}{*}{BNCI2014001}
& LaBraM
& 52.28 / 52.28
& 51.99 / 51.99
& 51.64 / 51.64
& \textbf{40.12 / 40.12}\\
& Neuro-GPT
& 54.88 / 54.88
& 54.46 / 54.46
& 50.44 / 50.44
& \textbf{40.82 / 40.82}\\
& CBraMod
& 53.65 / 53.65
& 53.74 / 53.74
& 53.41 / 53.41
& \textbf{46.60 / 46.60}\\

\cmidrule{1-6}
\multirow{3}{*}{SEED}
& LaBraM
& 55.61 / 55.37
& 51.38 / 51.07
& 49.10 / 48.69
& \textbf{45.44 / 45.07}\\
& Neuro-GPT
& 53.66 / 53.40
& 53.60 / 53.33
& 53.45 / 53.20
& \textbf{48.61 / 48.31}\\
& CBraMod
& 53.03 / 52.71
& 52.62 / 52.31
& 49.95 / 49.79
& \textbf{44.52 / 44.28}\\

\cmidrule{1-6}
\multirow{3}{*}{CHB-MIT}
& LaBraM
& 78.69 / 58.95
& 75.98 / 59.16
& 76.09 / 59.55
& \textbf{64.85 / 48.36}\\
& Neuro-GPT
& 80.54 / 68.46
& 80.45 / 68.38
& 82.06 / 67.70
& \textbf{76.55 / 63.33}\\
& CBraMod
& 68.75 / 58.80
& 68.44 / 58.64
& 69.18 / 58.73
& \textbf{64.29 / 53.62}\\

\midrule
\multicolumn{6}{c}{\textbf{Within-Subject}}\\
\addlinespace[2pt]
\multicolumn{6}{@{}l@{}}{\textbf{Linear Probing (LP)}}\\
\addlinespace[2pt]

\multirow{3}{*}{BNCI2014001}
& LaBraM
& 54.37 / 54.37
& 54.37 / 54.37
& 44.44 / 44.44
& \textbf{0.00 / 0.00}\\
& Neuro-GPT
& 51.59 / 51.59
& 50.79 / 50.79
& 38.89 / 38.89
& \textbf{9.92 / 9.92}\\
& CBraMod
& 50.20 / 50.20
& 50.00 / 50.00
& 41.47 / 41.47
& \textbf{12.90 / 12.90}\\

\cmidrule{1-6}
\multirow{3}{*}{SEED}
& LaBraM
& 78.40 / 78.25
& 61.22 / 60.96
& 35.14 / 35.07
& \textbf{6.15 / 6.17}\\
& Neuro-GPT
& 50.37 / 50.12
& 50.35 / 50.09
& 46.67 / 46.32
& \textbf{26.54 / 26.35}\\
& CBraMod
& 74.90 / 74.73
& 74.60 / 74.43
& 67.90 / 67.80
& \textbf{13.98 / 13.92}\\

\cmidrule{1-6}
\multirow{3}{*}{CHB-MIT}
& LaBraM
& 94.82 / 70.11
& 79.86 / 62.32
& 64.71 / 50.72
& \textbf{44.84 / 25.05}\\
& Neuro-GPT
& 82.20 / 70.46
& 82.29 / 70.27
& 78.19 / 67.41
& \textbf{50.15 / 37.80}\\
& CBraMod
& 92.49 / 70.52
& 92.53 / 70.37
& 91.21 / 67.69
& \textbf{76.67 / 47.79}\\

\midrule
\multicolumn{6}{@{}l@{}}{\textbf{Full Fine-Tuning (FT)}}\\
\addlinespace[2pt]

\multirow{3}{*}{BNCI2014001}
& LaBraM
& 49.80 / 49.80
& 49.01 / 49.01
& 43.45 / 43.45
& \textbf{2.98 / 2.98}\\
& Neuro-GPT
& 50.20 / 50.20
& 50.00 / 50.00
& 39.48 / 39.48
& \textbf{27.58 / 27.58}\\
& CBraMod
& 56.35 / 56.35
& 56.35 / 55.95
& 55.16 / 55.16
& \textbf{48.61 / 48.61}\\

\cmidrule{1-6}
\multirow{3}{*}{SEED}
& LaBraM
& 86.12 / 86.04
& 71.51 / 71.23
& 52.04 / 51.68
& \textbf{36.41 / 35.98}\\
& Neuro-GPT
& 73.86 / 73.65
& 73.77 / 73.54
& 71.46 / 71.22
& \textbf{64.25 / 63.93}\\
& CBraMod
& 87.09 / 87.00
& 85.68 / 85.59
& 82.08 / 82.00
& \textbf{69.78 / 69.63}\\

\cmidrule{1-6}
\multirow{3}{*}{CHB-MIT}
& LaBraM
& 74.87 / 64.02
& 75.53 / 63.49
& 75.42 / 61.11
& \textbf{74.41 / 59.01}\\
& Neuro-GPT
& 90.50 / 85.51
& 90.56 / 85.61
& 89.01 / 81.67
& \textbf{72.61 / 62.58}\\
& CBraMod
& 86.67 / 81.80
& 87.23 / 81.73
& 86.71 / 81.28
& \textbf{80.11 / 74.76}\\

\bottomrule
\end{tabular*}
\caption{Transfer attack results using projected gradient descent optimization under LOSO and within-subject evaluation. Each entry reports subject-averaged ACC/BACC (\%). TAA and SW-ProxyCE use the PGD-style optimizer. Lower attacked values indicate stronger attacks, and the strongest attacked result in each setting is shown in bold. GN denotes Gaussian noise, and Ours denotes SW-ProxyCE.}
\label{tab:appendix_pgd_acc_bacc}
\end{table*}

\section{H. Complete Analysis of Within-Class Variance Concentration}
\label{app:complete_cvr}

To complement the representative CVR analysis in the main paper, we reported the complete results across all three datasets and public encoders. The cumulative variance ratio (CVR) at \(k\) measures the fraction of observed within-class variance captured by the top \(k\) principal directions. The isotropic reference assigns equal variance to all observed directions and therefore represents an evenly distributed within-class spectrum.

\begin{figure*}[t]
\centering
\includegraphics[width=\textwidth]{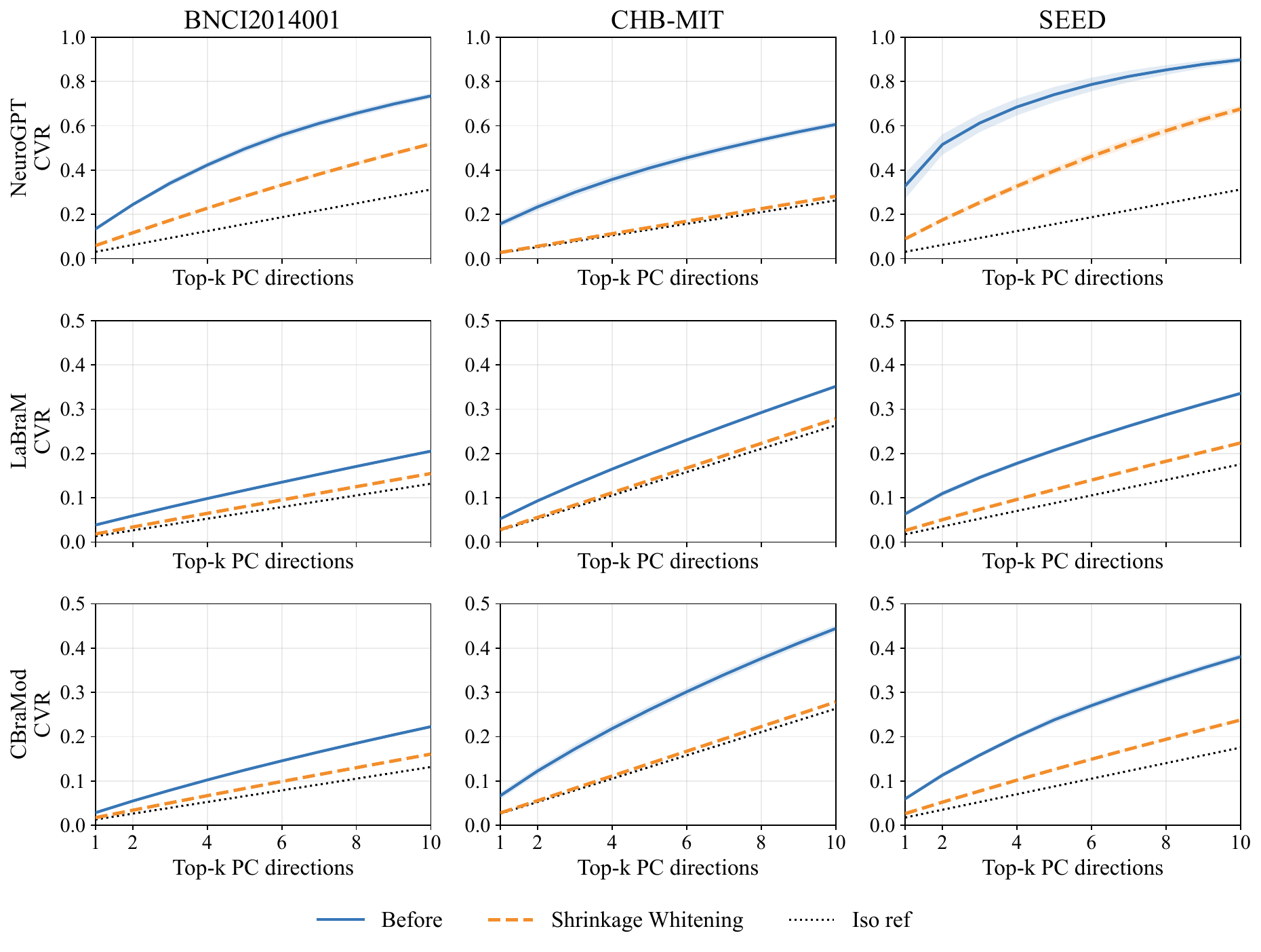}
\caption{Cumulative within-class variance ratio (CVR) across all dataset--encoder combinations using \(K=20\) references per class. Columns correspond to BNCI2014001, CHB-MIT, and SEED, while rows correspond to Neuro-GPT, LaBraM, and CBraMod. The solid blue curves show the variance spectra before whitening, the dashed orange curves show shrinkage whitening with \(\rho=0.4\), and the dotted black curves denote the isotropic reference. A lower CVR at a fixed \(k\) indicates less concentration in the leading within-class directions.}
\label{fig:complete_cvr}
\end{figure*}

As shown in Figure~\ref{fig:complete_cvr}, the spectra before whitening generally rise faster than the isotropic references, indicating that within-class variance is concentrated in a small number of leading directions. Shrinkage whitening consistently moves the CVR curves toward the isotropic references across the nine dataset--encoder combinations. Although the degree of anisotropy varies across datasets and encoders, the overall pattern confirms that SW-ProxyCE reduces the dominance of high-variance directions and produces a more balanced geometry for prototype construction and comparison.

\section{I. Qualitative Analysis of Adversarial Perturbations}
\label{app:qualitative_visualization}

\begin{figure*}[p]
\centering
\includegraphics[
    width=\textwidth,
    height=0.82\textheight,
    keepaspectratio
]{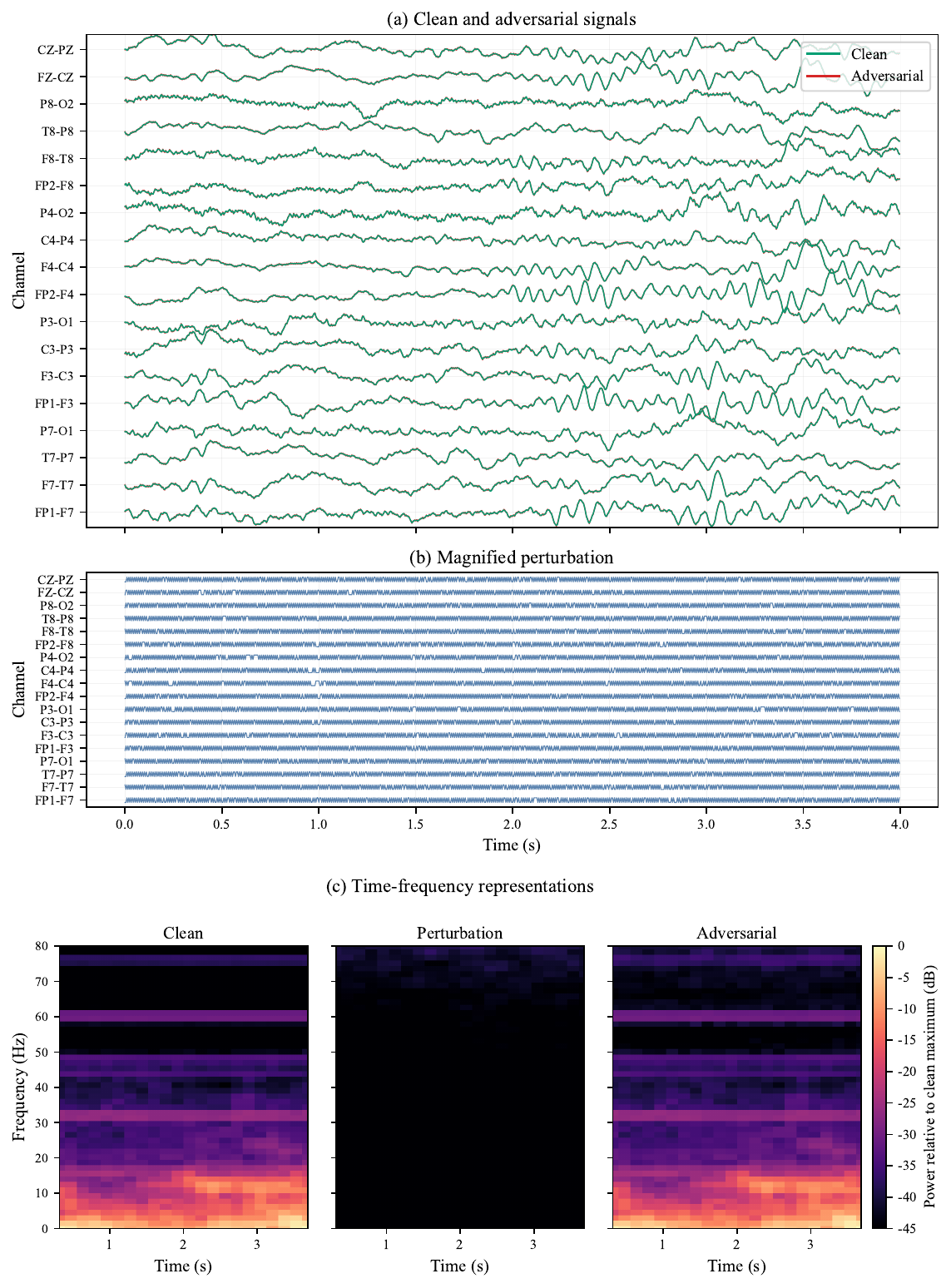}
\caption{Qualitative visualization of a successful SW-ProxyCE attack on CHB-MIT using a LaBraM-LP victim under LOSO evaluation. Panel (a) overlays the clean and adversarial EEG signals across all 18 bipolar channels, with the clean signals drawn above the adversarial signals. Panel (b) shows the corresponding perturbation magnified by \(8\times\). Panel (c) presents the mean-channel time--frequency representations of the clean input, perturbation, and adversarial input under a shared scale normalized by the maximum clean power. The attack uses \(K=20\), \(\epsilon=0.1\), changes the prediction from class 0 to class 1, and achieves an SNR of \(20.0\) dB for this trial.}
\label{fig:qualitative_labram}
\end{figure*}

\begin{figure*}[p]
\centering
\includegraphics[
    width=\textwidth,
    height=0.82\textheight,
    keepaspectratio
]{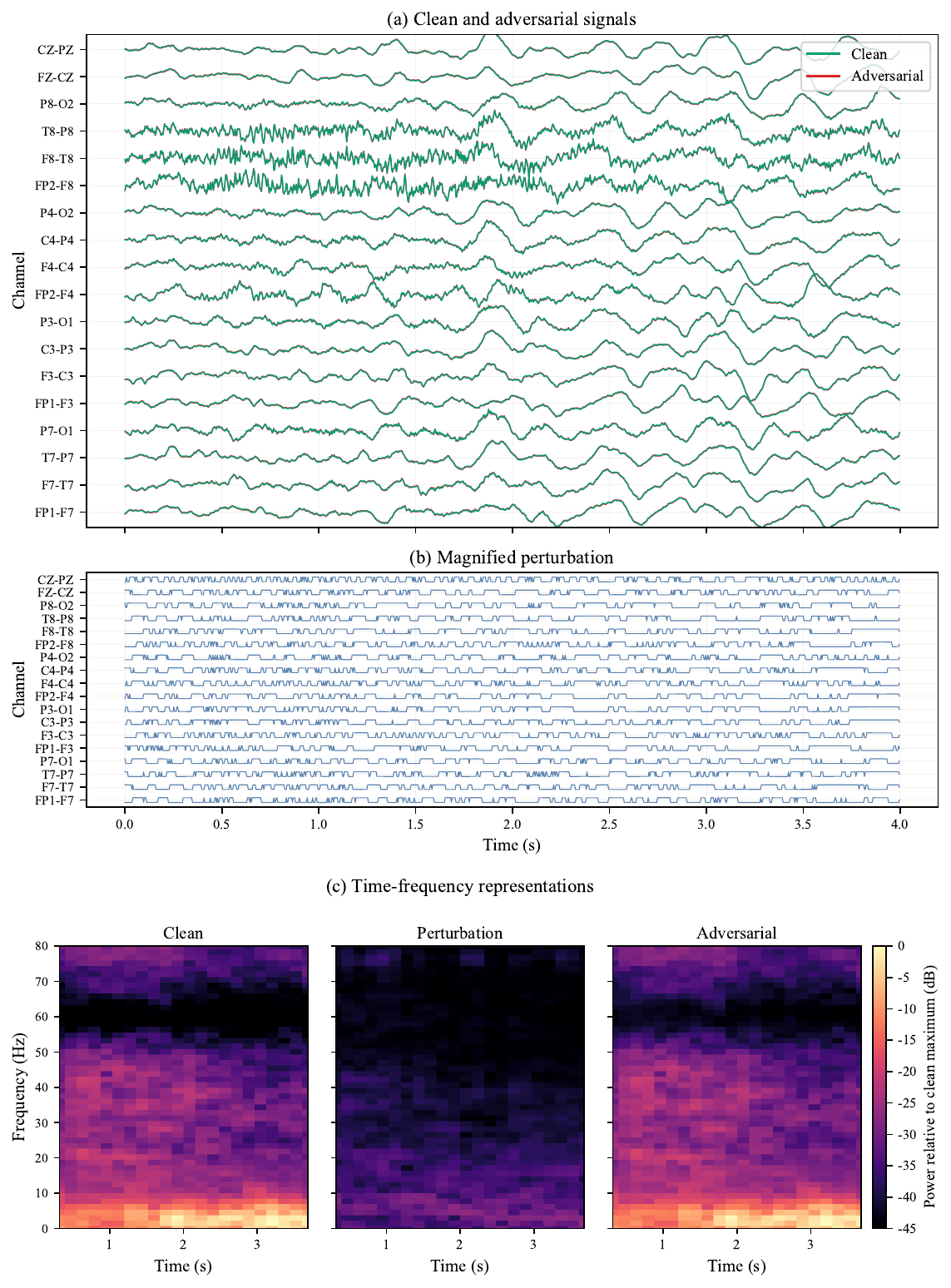}
\caption{Qualitative visualization of a successful SW-ProxyCE attack on CHB-MIT using a NeuroGPT-LP victim under LOSO evaluation. Panel (a) overlays the clean and adversarial EEG signals across all channels. Panel (b) shows the corresponding perturbation magnified by \(8\times\), and panel (c) compares the clean, perturbation, and adversarial time--frequency representations using a shared model-specific power scale. Although the perturbation distribution differs from that observed for LaBraM, the adversarial input retains the dominant temporal and spectral structures of the clean EEG. The attack uses \(K=20\), \(\epsilon=0.1\), changes the prediction from class 1 to class 0, and achieves an SNR of  \(20.0\) dB.}
\label{fig:qualitative_neurogpt}
\end{figure*}

\begin{figure*}[p]
\centering
\includegraphics[
    width=\textwidth,
    height=0.82\textheight,
    keepaspectratio
]{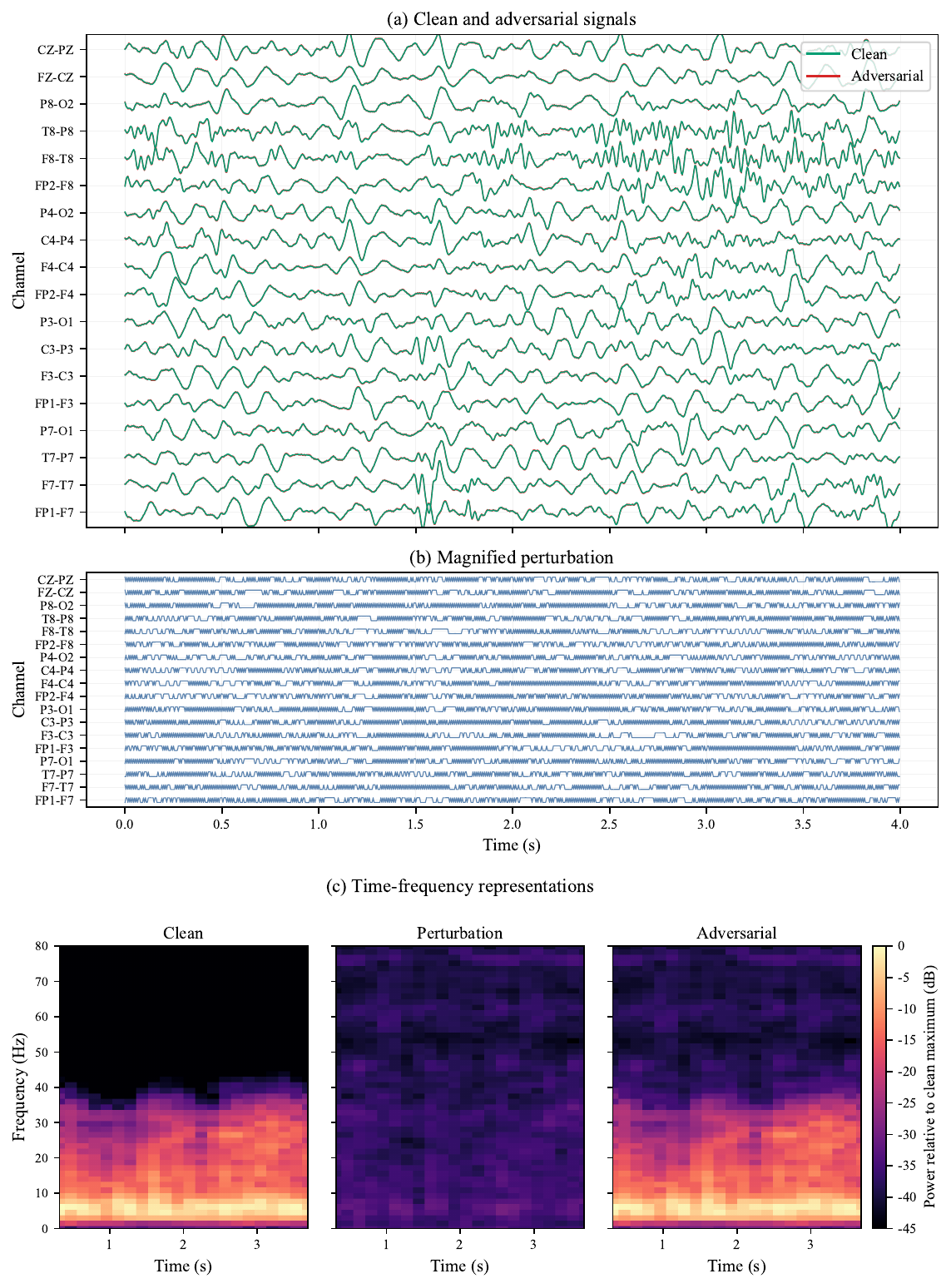}
\caption{Qualitative visualization of a successful SW-ProxyCE attack on CHB-MIT using a CBraMod-LP victim under LOSO evaluation. Panels (a) and (b) show the clean/adversarial waveform overlay and the \(8\times\)-magnified perturbation, respectively. Panel (c) shows the corresponding mean-channel time--frequency representations under a shared power scale normalized by the clean input. The adversarial input preserves the principal time-domain morphology and dominant time--frequency content despite the successful prediction change. The attack uses \(K=20\), \(\epsilon=0.1\), changes the prediction from class 1 to class 0, and achieves an SNR of \(20.0\) dB.}
\label{fig:qualitative_cbramod}
\end{figure*}

We further examined representative successful adversarial examples to characterize the input-space changes introduced by SW-ProxyCE beyond the aggregate attack performance reported in the main paper. We considered both time-domain waveforms and time--frequency representations, which provide complementary views of the perturbation magnitude and its effect on the underlying EEG structure. To examine whether the observed characteristics are specific to one encoder, we presented separate visualizations for LaBraM, NeuroGPT, and CBraMod in Figures~\ref{fig:qualitative_labram}, \ref{fig:qualitative_neurogpt}, and \ref{fig:qualitative_cbramod}, respectively. All three examples are successful attacks on CHB-MIT under LOSO-LP evaluation with \(K=20\) references per class and an \(\ell_\infty\) budget of \(\epsilon=0.1\).

\paragraph{Time-domain characteristics.}
Panels (a) and (b) of Figures~\ref{fig:qualitative_labram}--\ref{fig:qualitative_cbramod} show the time-domain signals and the corresponding adversarial perturbations. For visualization, each channel is centered and scaled using the temporal statistics of its corresponding clean channel, after which the same transformation is applied to the adversarial signal. The transformed channels are then vertically offset to permit simultaneous inspection of all 18 bipolar channels. This display transformation is used only for visualization and does not alter the adversarial examples evaluated by the victim models.

Across the three encoders, the clean and adversarial signals almost completely overlap at the original display scale. The examples do not exhibit conspicuous amplitude shifts, abrupt discontinuities, or large-scale deformations of the original waveforms. The red adversarial traces are largely covered by the green clean traces, and their local differences become clearly visible only after the perturbations are magnified by \(8\times\) in panel (b). Thus, the magnified views confirm that nonzero and temporally structured perturbations are present, while panel (a) shows that these perturbations remain visually subtle at the original scale.

Despite the strong waveform overlap, all three perturbations successfully alter the corresponding downstream predictions. The LaBraM example changes the victim prediction from class 0 to class 1, whereas the NeuroGPT and CBraMod examples change the prediction from class 1 to class 0. Their measured SNRs are \(20.0\) dB, and their maximum absolute perturbations satisfy the specified \(\ell_\infty\) budget of \(0.1\). These examples demonstrate that a successful SW-ProxyCE attack does not require visually prominent corruption of the complete EEG waveform. Instead, relatively small input modifications can exploit directions to which the public encoder and its downstream victim remain sensitive.

\paragraph{Time--frequency characteristics.}
Panel (c) of each figure further compares the mean-channel time--frequency representations of the clean input, adversarial perturbation, and adversarial input. The representations are computed from the corresponding model-specific inputs before the additional display transformation used in panels (a) and (b). Within each figure, the three time--frequency panels share the same power scale, normalized by the maximum power of the corresponding clean input. Consequently, the perturbation energy can be directly compared with both the original spectral structure and the structure retained in the adversarial example.

For all three encoders, the adversarial input closely preserves the dominant time--frequency pattern of the corresponding clean EEG. The principal low-frequency energy distribution, its temporal evolution, and the major spectral structures remain visually consistent after the attack. In contrast, the perturbation panels generally exhibit substantially lower power under the shared model-specific scale. This observation is particularly apparent for LaBraM, for which the perturbation energy is weak relative to the dominant clean spectral components. NeuroGPT and CBraMod exhibit broader perturbation distributions, but their adversarial inputs still retain the main time--frequency structures visible in the clean inputs.

The detailed perturbation patterns differ across the three models. Such differences are expected because LaBraM, NeuroGPT, and CBraMod use different encoder architectures, model-specific input sampling rates, and different representation mappings. SW-ProxyCE optimizes the perturbation through each public encoder, and therefore the resulting input-space direction need not have an identical temporal or spectral form across models. Importantly, the shared observation is not that the perturbations are spectrally identical, but that each model can be attacked without broadly reconstructing or replacing the dominant time--frequency content of its clean EEG input.

Taken together, the model-wise visualizations show that SW-ProxyCE can change downstream predictions while introducing only subtle modifications relative to the original EEG. The time-domain plots show strong overlap between clean and adversarial waveforms, while the time--frequency plots show that the principal spectral structures are retained after attack. These qualitative examples complement the dataset-level BACC, ACC, and ASR results by demonstrating that the observed performance degradation is not accompanied by visually prominent waveform distortion or wholesale alteration of the dominant spectral content.

%\clearpage

%\clearpage

\end{document}